\documentclass[acmtog, authorversion, nonacm]{acmart}

\AtBeginDocument{%
  \providecommand\BibTeX{{%
    \normalfont B\kern-0.5em{\scshape i\kern-0.25em b}\kern-0.8em\TeX}}}

\acmJournal{FACMP}
\usepackage{booktabs} %

\usepackage[ruled]{algorithm2e} %
\usepackage{float}
\usepackage{caption}
\usepackage{subcaption}
\usepackage{placeins}

\SetAlFnt{\small}
\SetAlCapFnt{\small}
\SetAlCapNameFnt{\small}
\SetAlCapHSkip{0pt}

\acmJournal{TOG}

\usepackage{amsmath}
\usepackage{comment}
\usepackage{multirow,bigdelim}
\usepackage{lipsum}
\usepackage{array}
\usepackage{wrapfig}

\usepackage{adjustbox}
\usepackage[percent]{overpic}
\usepackage{makecell}
\usepackage[normalem]{ulem}
\usepackage{blindtext}
\usepackage{xcolor}
\usepackage{soul}
\usepackage{cleveref}
\usepackage{amsfonts}
\usepackage{subcaption}

\makeatletter
\newcommand{\settitle}{\@maketitle}
\makeatother

\newcolumntype{C}[1]{>{\centering\let\newline\\\arraybackslash\hspace{0pt}}m{#1}}

\newif\ifdraft
\drafttrue

\ifdraft
\definecolor{darkpink}{rgb}{0.561, 0.282, 0.427}
\definecolor{darkturquoise}{rgb}{0., 0.81, 0.822}

\newcommand{\dcc}[1]{{\color{red}[\textbf{DC:} #1]}}
\newcommand{\rgc}[1]{{\color{purple}[\textbf{RG:} #1]}}
\newcommand{\opc}[1]{{\color{blue}[\textbf{OP:} #1]}}

\newcommand{\abc}[1]{{\color{green}[\textbf{AB:} #1]}}

\newcommand{\drop}[1]{}

\else
\newcommand{\dcc}[1]{}
\newcommand{\rgc}[1]{}
\newcommand{\opc}[1]{}
\newcommand{\gcc}[1]{}
\newcommand{\hmc}[1]{}
\newcommand{\abc}[1]{}

\fi

\def\Naive{Na\"{\i}ve\xspace}

\makeatletter
\DeclareRobustCommand\onedot{\futurelet\@let@token\@onedot}
\def\@onedot{\ifx\@let@token.\else.\null\fi\xspace}

\def\eg{\emph{e.g}\onedot}

\def\pholdercolor{{\color{blue}$\left[S_*\right]$}}

\makeatother

\usepackage[bottom]{footmisc}
\usepackage{arydshln}

\makeatletter
\def\blfootnote{\xdef\@thefnmark{}\@footnotetext}
\makeatother

\begin{document}
\title{Domain-Agnostic Tuning-Encoder for Fast Personalization of Text-To-Image Models}

\author{Moab Arar}
\affiliation{%
 \institution{Tel Aviv University}
 \city{Tel Aviv}
 \country{Israel}}

 \author{Rinon Gal}
\affiliation{%
 \institution{Tel Aviv University, NVIDIA}
 \city{Tel Aviv}
 \country{Israel}}
\authornote{
Work was done during an internship at NVIDIA
}

\author{Yuval Atzmon}
\affiliation{%
 \institution{NVIDIA}
 \city{Tel Aviv}
 \country{Israel}}
 
\author{Gal Chechik}
\affiliation{%
 \institution{NVIDIA}
 \city{Tel Aviv}
 \country{Israel}}
 
\author{Daniel Cohen-Or}
\affiliation{%
 \institution{Tel Aviv University}
 \city{Tel Aviv}
 \country{Israel}}

\author{Ariel Shamir}
\affiliation{%
 \institution{Reichman University (IDC)}
 \city{Herzliya}
 \country{Israel}}
 
 \author{Amit H. Bermano}
\affiliation{%
 \institution{Tel Aviv University}
 \city{Tel Aviv}
 \country{Israel}}

\begin{abstract}

Text-to-image (T2I) personalization allows users to guide the creative image generation process by combining their own visual concepts in natural language prompts. Recently, encoder-based techniques have emerged as a new effective approach for T2I personalization, reducing the need for multiple images and long training times. However, most existing encoders are limited to a single-class domain, which hinders their ability to handle diverse concepts. In this work, we propose a domain-agnostic method that does not require any specialized dataset or prior information about the personalized concepts. We introduce a novel contrastive-based regularization technique to maintain high fidelity to the target concept characteristics while keeping the predicted embeddings close to editable regions of the latent space, by pushing the predicted tokens toward their nearest existing CLIP tokens. Our experimental results demonstrate the effectiveness of our approach and show how the learned tokens are more semantic than tokens predicted by unregularized models. This leads to a better representation that achieves state-of-the-art performance while being more flexible than previous methods~\footnote{Our project page available at~\url{https://datencoder.github.io}}. 
\end{abstract}

\begin{teaserfigure}
    \centering
     \includegraphics[width=0.98\linewidth]{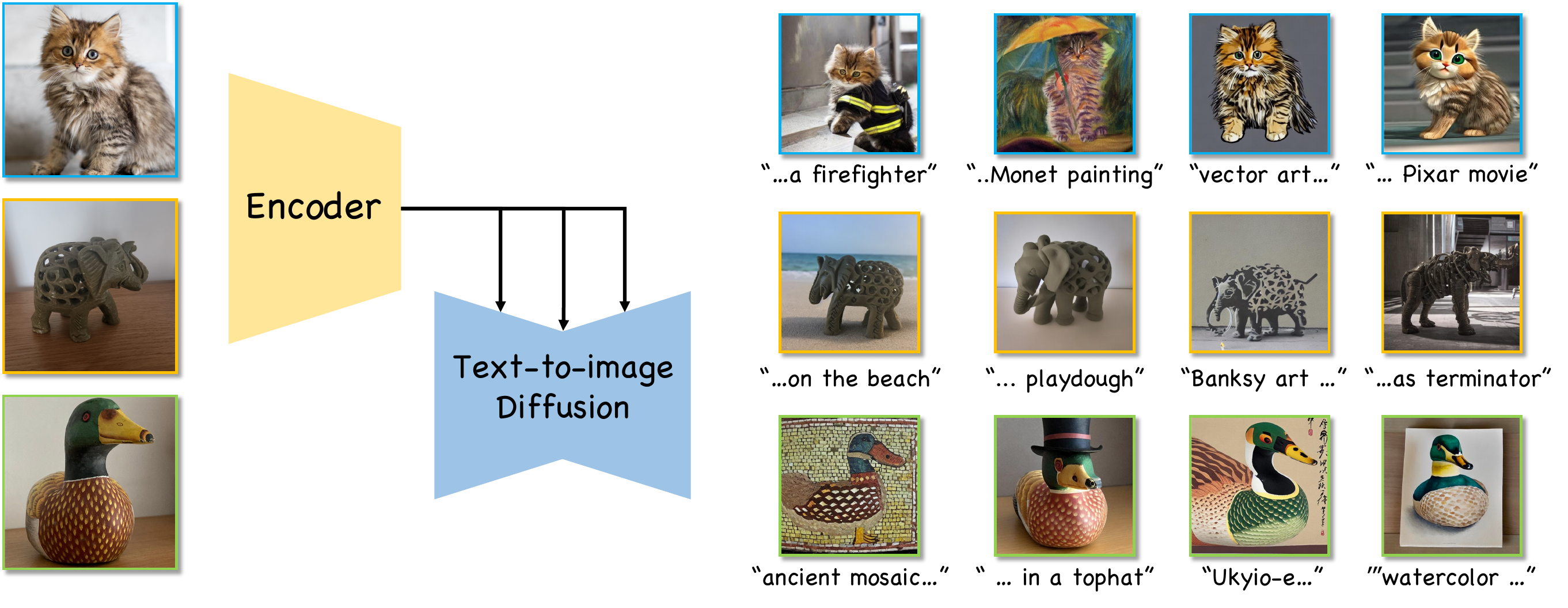}
    \caption{Our domain-agnostic tuning-encoder can personalize a text-to-image diffusion model to a given concept using $12$ or fewer training steps, allowing for general one-shot inference-time tuning. The personalized models are used to generate images of the concept in new settings using natural language prompts.}
    \label{fig:teaser}
\end{teaserfigure}
\maketitle
\section{Introduction}

The rapid advancement of generative models has revolutionized content creation, enabling effortless generation of diverse artworks. Part of their true potential lies in personalization, allowing users to tailor outputs to unique personal concepts. Personalizing a model involves customizing it to capture and manifest unique characteristics of personal belongings, memories, or self-portraits. However, early personalization methods~\citep{gal2022image,ruiz2022dreambooth} rely on the availability of multiple images and require lengthy optimization.

An effective alternative is pre-training predictive models for targeting concepts. These approaches train an encoder to predict a text embedding that accurately reconstructs a given desired target concept. Using the obtained embeddings, one can generate scenes portraying the given concept. Still, such methods face limitations. First, they rely on a single-class domain, which constrains their ability to capture the long tail distribution of diverse concepts. Second, some approaches necessitate external priors, such as segmentation masks or multi-view input, to effectively capture the characteristics of the target concept while discarding spurious background features.

In this work, we follow E4T~\citep{gal2023designing}, an approach which leverages the encoder as a form of initialization for brief ($5$-$15$ iteration) fine-tuning. E4T trains an encoder for each individual domain, and requires roughly 70GB of VRAM for inference-time tuning. Our approach can tackle multiple domains, and reduces inference-time memory requirements. We consider two goals while designing our encoder: (1) the ability to edit the target concepts, and (2) the ability to faithfully capture distinguishable characteristics of the target. We achieve the first goal by regularizing the model to predict words within the editable region of the generative model. Unlike prior single-domain methods, we do not rely on a coarse description of the target domain. Instead, we use a contrastive-based approach to push the predicted embedding toward meaningful regions in the word embedding space. Intuitively, ensuring the prediction is near words that semantically describe the concept class will better preserve the model's prior knowledge of the concept class. 

For the second goal, we introduce a hyper-network to capture the distinctive features of the target concepts with higher fidelity. To ensure a manageable model size, we employ a strategy of predicting a low-rank decomposition of the weights of the UNET-denoiser model, following the approach outlined in \citet{Hu2021LoRALA} and \citet{gal2023designing}. 
Finally, the joint embedding and hyper-network predictions are used to initialize a regularized LoRA training process, requiring $12$ or fewer optimization steps. Importantly, this reduces memory requirements from roughly 70GB to fewer than 30GB and shortens training and inference times.

We compare our method to existing encoders and optimization-based approaches and demonstrate that it can achieve high quality and fast personalization across many different domains.

\section{Related work}
\begin{figure*}
    \centering

    \includegraphics[width=0.98\textwidth]{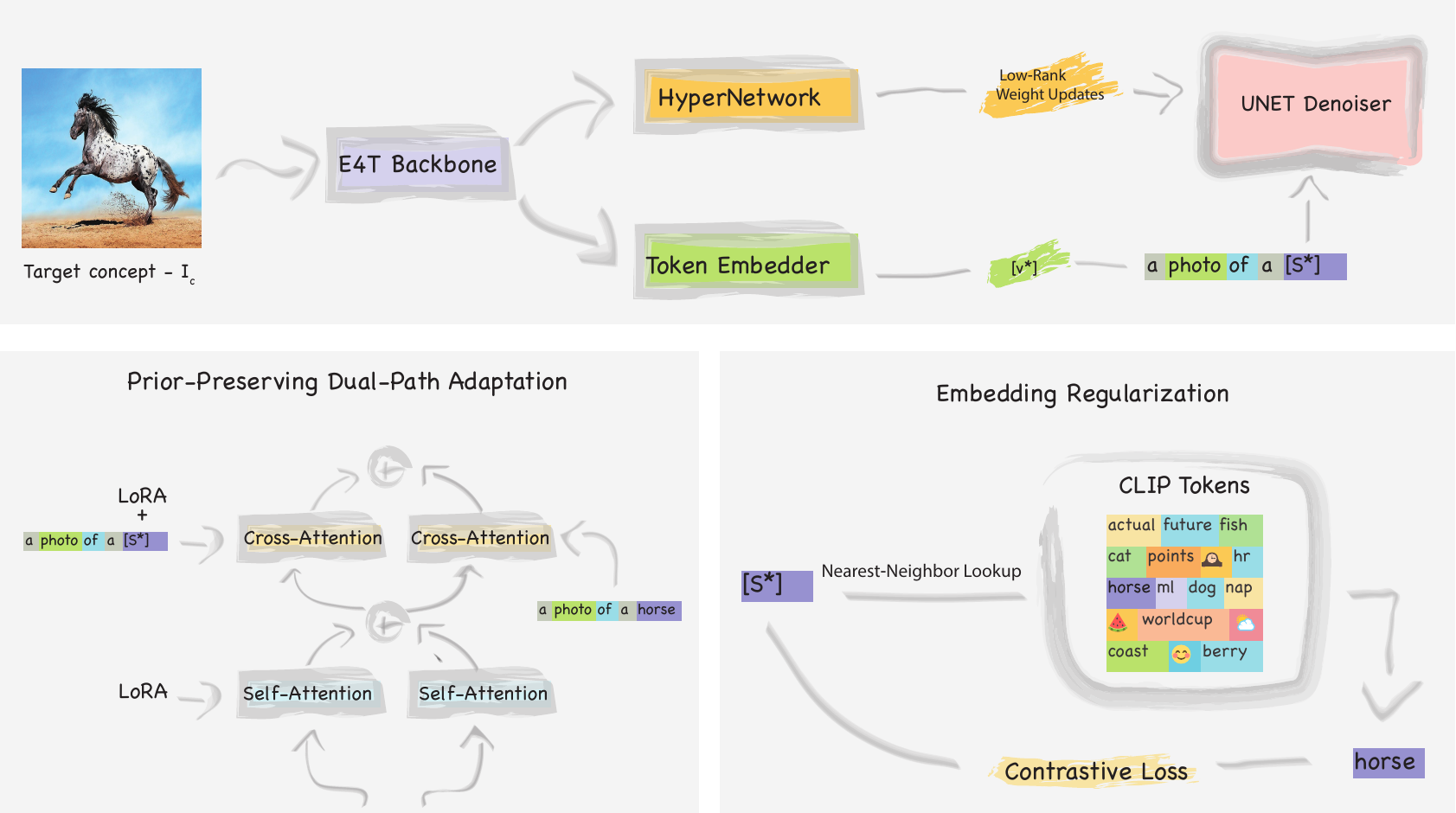}
    \caption{\textbf{Method overview.} (top) Our method consists of a feature-extraction backbone which follows the E4T approach and uses a mix of CLIP-features from the concept image, and denoiser-based features from the current noisy generation. These features are fed into an embedding prediction head, and a hypernetwork which predicts LoRA-style attention-weight offsets. (bottom, right) Our embeddings are regularized by using a nearest-neighbour based contrastive loss that pushes them towards real words, but away from the embeddings of other concepts. (bottom, left) We employ a dual-path adaptation approach where each attention branch is repeated twice, once using the soft-embedding and the hypernetwork offsets, and once with the vanilla model and a hard-prompt containing the embedding's nearest neighbor. These branches are linearly blended to better preserve the prior.}
    \label{fig:arch-fig}
\end{figure*}

\paragraph{\textbf{Text-driven image generation using diffusion models.}}
Text-to-image synthesis has made significant progress in recent years, driven mainly by pre-trained diffusion models~\citep{ho2020denoising} and especially by large models~\citep{balaji2022ediffi,rombach2021highresolutionLDM,ramesh2022hierarchical,nichol2021glide} trained on web-scale data like ~\citep{schuhmann2021laion}.
Our approach builds upon these pre-trained models to extend their vocabulary and generate personalized concepts. Specifically, we use the Stable-Diffusion model \citep{rombach2021highresolutionLDM}. We expect it to generalize to diffusion-based generators with similar attention-based architectures \citep{saharia2022photorealistic}.

\paragraph{\textbf{Text-based image editing}} Following the success of CLIP-based \citep{radford2021learning} editing methods~\citep{patashnik2021styleclip,gal2021stylegan,text2mesh,bar2022text2live}, a large body of work sought to leverage the power of recent large-scale text-to-image models~\citep{rombach2021highresolutionLDM,ramesh2022hierarchical,saharia2022photorealistic,balaji2022ediffi,Sauer2023ICML,kang2023gigagan} in order to manipulate images using text-based guidance. Prompt-to-Prompt~\citep{hertz2022prompt} propose a method for manipulating \emph{generated} images by re-using an initial prompt's attention masks. In a follow-up work, \citet{mokady2022null} extend this approach to real images by encoding them into the null-conditioning space of classifier-free guidance~\citep{ho2021classifier}. \citet{tumanyan2022plug} and \citet{parmar2023zeroshot} extract reference attention maps or features using DDIM~\citep{song2020denoising} based reconstructions. These are then used to preserve image structure under new prompts. Others train an instruction-guided image-to-image translation network using synthetic data~\citep{brooks2022instructpix2pix} or tune the model to reconstruct an image and use conditioning-space walks to modify it~\citep{kawar2022imagic}. Such approaches can also be used to edit 3D content, for example modifying shapes textures using depth and text guided diffusion models~\citep{richardson2023texture}.

Common to these image-editing approaches is a desire to preserve the content of the original image. In contrast, our method deals with model personalization which aims to capture a concept for later use in novel scenes.
There, the aim is to learn the semantics and apperance of a subject, but not its specific structure in the image.

\paragraph{\textbf{Inversion.}} In the context of Generative Adversarial Networks (GANs, \citep{goodfellow2014generative}), inversion is the task of finding a latent representation that will reproduce a specific image when passed through a pre-trained generator~\citep{zhu2016generative,xia2021gan}. There, methods are split into two main camps. In the first are optimization methods, which iterative search the latent space for a code that synthesizes an image with some minimal reconstruction loss~\citep{abdal2019image2stylegan,abdal2020image2stylegan++,zhu2020improved,gu2020image}. In the second are encoder based methods, which train a neural network to directly predict such latents~\citep{richardson2020encoding,zhu2020domain,pidhorskyi2020adversarial,tov2021designing,wang2021HFGI,bai2022high,parmar2022spatially}.

With diffusion models, the inversion latent space can be the initial noise map that will later be denoised into a given target~\citep{dhariwal2021diffusionBeatsGAN,ramesh2022hierarchical,song2020denoising}. In a more recent line of work, inversion has been used to refer to finding a conditioning code that can be used to synthesize novel images of a given concept~\citep{gal2022image}. There, the goal is not to recreate a specific image, but to capture the semantics of a concept outlined in one or more target images and later re-create it in new scenes. Our approach similarly aims to encode a concept.

\paragraph{\textbf{Personalization.}} 
Personalization methods aim to tune a model to a specific individual target. Often, the goal is to combine some large-scale prior knowledge with unique information associated with an end-user. These can include personalized recommendation systems~\citep{benhamdi2017personalized,fernando2018artwork,martinez2009s,cho2002personalized}, federated learning~\citep{mansour2020three,jiang2019improving,fallah2020personalized,shamsian2021personalized}, or the creation of generative models tuned on specific scenes or individuals~\citep{semantic2019bau,roich2021pivotal,alaluf2021hyperstyle,dinh2022hyperinverter,cao2022authentic,nitzan2022mystyle,cohen2022my}. In text-to-image personalization, the goal is to teach pre-trained models to synthesize novel images of a specific target concept, guided by natural language prompts. Initial work in this field employed direct optimization approaches, either tuning a set of text embeddings to describe the concept~\citep{gal2022image, voynov2023p+}, modifying the denoising network itself~\citep{ruiz2022dreambooth,han2023svdiff}, or a mixture of both~\citep{kumari2022multi,tewel2023key,simoLoRA2023}. However, such optimization-based approaches require lengthy training sessions, typically requiring dozens of minutes for every concept. 

More recently, encoder-based approaches emerged~\cite{wei2023elite,shi2023instantbooth,gal2023designing,zhou2023enhancing,li2023blipdiffusion}, which train a neural network to predict some latent representation that can be injected into the network to synthesize new images of the concept. These either require subject-specific segmentation masks~\citep{wei2023elite} or use single-domain training to both regularize the model and allow it to infer the target from the single image~\citep{gal2023designing,shi2023instantbooth}. In an alternative approach, a model can be trained to synthesize novel images from dual conditions: a text prompt, and a set of images depicting the target~\citep{Chen2023SubjectdrivenTG}. However, this approach is based on apprenticehsip learning, where the model is trained on outputs from half a million pre-trained personalized models. Such an approach therefore requires roughly $14$ A100 GPU-years, making it infeasible for most practitioners.

Our method follows the encoder-based approach, but extends it beyond the single-domain without use of any segmentation masks or additional labels. Moreover, compared to prior encoder-based tuning approaches~\citep{gal2023designing}, our tuning-phase is quicker and has reduced memory overhead.

\section{Preliminaries}

To put our contribution in context, we begin with an overview of two recent text-to-image personalization approaches: Textual Inversion~\citep{gal2022image} and E4T~\citep{gal2023designing} which serve as a basis for our work. 

\subsection{Textual Inversion}
\label{sec:method_ti}
Textual Inversion (TI) introduced the topic of text-to-image (T2I) personalization, where a pre-trained T2I diffusion model is taught how to reason about unique, user-provided concepts which were unseen during training. In TI, the authors propose to tackle this task by learning a novel word-embedding, $v*$, that will represent a concept visualized in a small ($3$-$5$) image set. To find such an embedding, the authors leverage the simple diffusion denoising loss~\citep{ho2020denoisingDDPM}:
\begin{equation}
    L_{Diffusion} := \mathbb{E}_{z, y, \epsilon \sim \mathcal{N}(0, 1), t }\Big[ \Vert \epsilon - \epsilon_\theta(z_{t},t,y) \Vert_{2}^{2}\Big] \, ,
    \label{eq:l_simple}
\end{equation}
where $\epsilon$ is the unscaled noise sample, $\epsilon_\theta$ is the denoising network, $t$ is the time step, $z_t$ is an image or latent noised to time $t$, and $c$ is some conditioning prompt containing an arbitrary string $S_*$ that is mapped to the embedding $v*$.

Once learned, this embedding can be invoked in future prompts (by including the placeholder $S_*$, \eg ``a photo of $S_*$") in order to generate images of the concept in novel contexts and scenes.

\subsection{Encoder for Tuning (E4T): } 
\label{sec:method_ti}
Although optimization-based approaches like TI can reconstruct the target concept, they require many iterations to converge. Indeed, personalizing a model with TI typically requires dozens of minutes even on commercial-grade GPUs. Recently, encoder-based approaches have emerged that train a neural network to directly map an image of a concept to a novel embedding. More concretely, given an input image $I_c$ depicting the concept, the encoder $E$ is trained to predict a suitable embedding: $v_* = E(I ; \theta)$. This encoder can be pretrained on a large set of images using the same denoising goal of \cref{eq:l_simple}, allowing it to later generalize to new concepts.

In E4T, this encoder is pre-trained on a single target domain (\eg human faces, cats or artistic styles). However, in order to prevent overfitting and preserve editability, it regularizes the predicted embeddings by restricting them to a region close to the embedding of a word describing the single domain (\eg ``face", "cat" or "art"). This regularization comes at the cost of identity preservation, which the authors later restore through an inference-time tuning session using a single target image of the concept and a few seconds of training.

Our goal is to extend this encoder-based tuning approach to an unrestricted domain, allowing a user to quickly personalize a model even for rare concepts for which large training sets may not exist.

\section{Method}

\subsection{Architecture Desgin}

We adopt the E4T architecture, which features an iterative-refinement design. Specifically, we utilize a pre-trained CLIP~\citep{radford2021learning} ViT-H-14 visual encoder and StableDiffusion's UNET-Encoder as feature-extraction backbones. We extract the spatial features for the given input image from each backbone's last layer. Following E4T, when extracting features from the UNET-Encoder, we provide it with an empty prompt. The features are processed by a convolutional-based network and shared between two prediction heads: a token embedder and a HyperNetwork. The token embedder predicts word embeddings that will be used to represent our target concept $I_c$. The HyperNetwork predicts weight-modulations for Stable Diffusion's denoising UNET. Next we discuss some important aspects about each prediction head.

\paragraph{\textbf{HyperNetwork: } } It is challenging to capture the fine details of the target concept by using only a token embedding. Previous works showed that modulating subsets of the denoiser weights can improve reconstruction quality with minor harm to the model's prior. Therefore, we seek to predict a set of weight modulations to help tune the denoiser for better identity preservation. Moreover, we make use of Stable Diffusion~\citep{rombach2021highresolutionLDM}, which consists of roughly a billion parameters. Adapting so many weights using a HyperNetwork is computationally infeasible. Hence, we follow prior art~\citep{gal2023designing,kumari2022multi,simoLoRA2023} and focus on predicting modulations for a subset of Stable Diffusion's layers, and specifically for the attention projection matrices. However, Stable Diffusion contains $96$ such matrices, each containing an average of $715,946$ parameters. Predicting such large matrices is still challenging. Instead, we predict decomposed weights of the same form as Low-Rank Adaptation (LoRA)~\cite{Hu2021LoRALA}, where each weight, $W \in \mathbb{R} ^{D_{in} \times D_{out}}$, is modulated by injecting trainable rank decomposition matrices. More specifically, for each concept $I_c$ and each projection matrix W, we predict two matrices , $A \in \mathbb{R} ^{D_{in} \times r}$ and $B \in \mathbb{R} ^{r \times D_{out}}$, where $r$ is the decomposition rank. The the new modulated matrices are:

\begin{equation}
W'= W + \Delta W = W+A \times B
\end{equation}.

To avoid breaking the model at the beginning of training, we initialize the prediction layer of the matrix $B$ to zero, and scale $\Delta W$ by a constant factor following~\cite{Hu2021LoRALA}. We further regularize the weight-offsets by applying $L2$-regularization.

\subsection{Embedding Regularization}
Large Language models are trained on a finite dictionary composed of tokens. Particularly, these models process words by dividing them into a sequence of tokens from the dictionary, which are then converted into appropriate embeddings $\{T_{i}\}_{i=1}^{n}$. In this tokenization process, each word is mapped to one or more high-dimensional vectors, which is used as input for transformer-based model. 

Our encoder's objective is to predict an embedding, $v* = E(I_c)$, that best describes a target-concept $I_c$. Previous works~\citep{gal2023designing} have shown that in under-constrained settings, encoders tend to use out-of-distribution embeddings. These tend to draw attention away from other words~\citep{tewel2023key}, limiting the ability to later manipulate the personalized concept via novel prompts. To prevent this attention-overfitting, we could use existing token embeddings to describe $I_c$. While these tokens are within the training distribution and hence editable, they are not expressive enough to capture personal concepts. We thus relax this hard constraint and predict embeddings \textit{close} to existing tokens. Intuitively, constraining $E(I_c)$ near semantically related words balances the trade-off between reconstruction and editing. However, unlike in single-domain encoders, where a coarse description of the domain exists and is known a-priori, in our setting there could be many semantically different words describing different concepts in the training data. Moreover, the domain encountered during inference may differ from those observed in training.

Inspired by~\cite{huang2023reversion, miech2020endcontrastive}, we make use of a "nearest-neighbor" contrastive-learning objective with dual goals: (1) push the predicted embedding close to their nearest CLIP tokens, and (2) map different concept images to different embeddings. Concretely, given $v_{*}=E(I_c)$, we find $\mathbb{N} \left( v_{*} \right)$, the set of nearest CLIP-tokens to $v_{*}$ in terms of the cosine distance metric. These CLIP tokens, $T_i \in \mathbb{N} \left( v_{*} \right)$ serve as positive examples in the contrastive loss. For every other image $I' \neq I_c$ in the current mini-batch, we use the embedding $v'=E(I')$ as our negative sample. Therefore, our loss is defined by:

\begin{equation}
    L_{c}(v_{*}) = - \log \frac{ \sum_{\mathbb{N} \left( v_{*} \right)} \exp \left( {v_* \cdot T_i } / {\tau} \right) }{  \sum_{ \mathbb{N} \left( v_{*} \right)} \exp \left( v_* \cdot T_i / \tau \right) +  \sum_{ v' \neq v_{*}} \exp \left( v_* \cdot v' / \tau \right) } 
\end{equation}
. As opposed to previous methods~\cite{huang2023reversion}, using the nearest neighbors embeddings as positive samples requires no supervision or prior knowledge on the target domain, canceling the need for a pre-defined list of positive and negative tokens in advance. Finally, we additionally employ an L2-regularization term to prevent the norm of the embeddings from increasing significantly:
\begin{equation}
    L_{L2}(v_{*}) = ||v_{*}||^2
\end{equation}
\begin{figure*}[!htb]

    \centering
    \setlength{\tabcolsep}{1.5pt}
    {\small
    \begin{tabular}{c c c c c c c c c c}

        \multirow{2}{*}{Input} & \multirow{2}{*}{Prompt} & Textual Inversion & DreamBooth & \multicolumn{2}{c}{LoRA} & ELITE & Ours\\

         &  & (Few-shot) & (Few-shot) & (One-shot) & (Few-shot) & ( w/ Seg Mask) & (One-shot) \\

        \includegraphics[width=0.117\textwidth,height=0.117\textwidth]{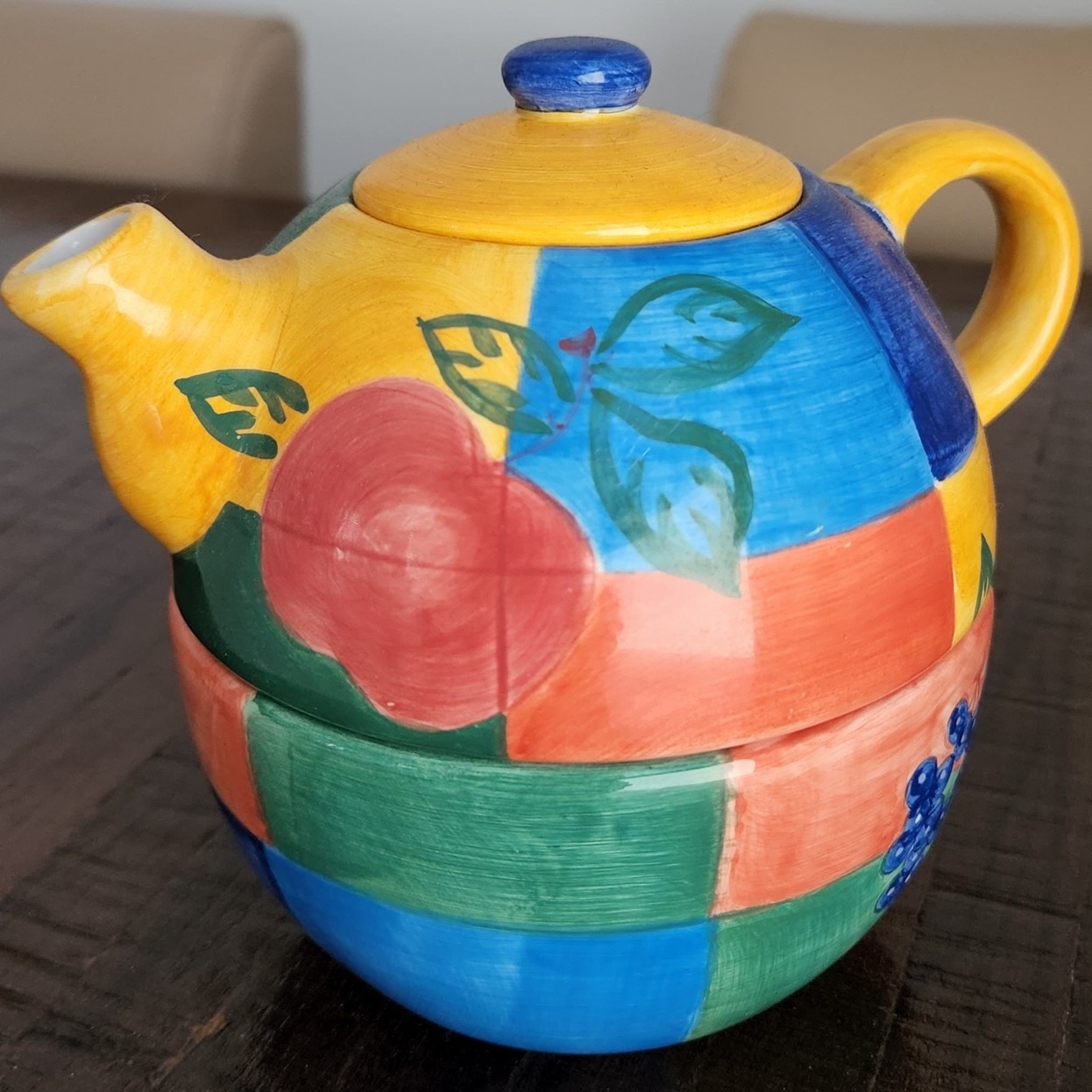} & \raisebox{0.048\textwidth}{\rotatebox[origin=t]{0}{\scalebox{0.9}{\begin{tabular}{c@{}c@{}c@{}} \pholdercolor{} themed backpack\end{tabular}}}} &
        \includegraphics[width=0.117\textwidth,height=0.117\textwidth]{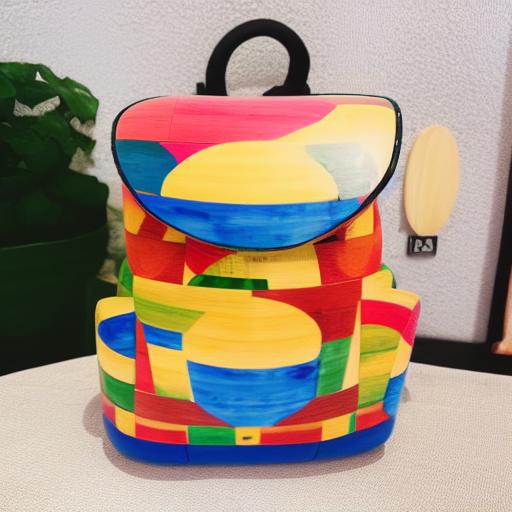} &
        \includegraphics[width=0.117\textwidth,height=0.117\textwidth]{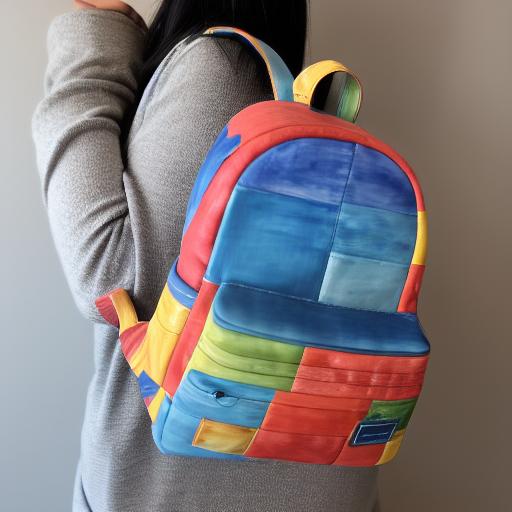} &
        \includegraphics[width=0.117\textwidth,height=0.117\textwidth]{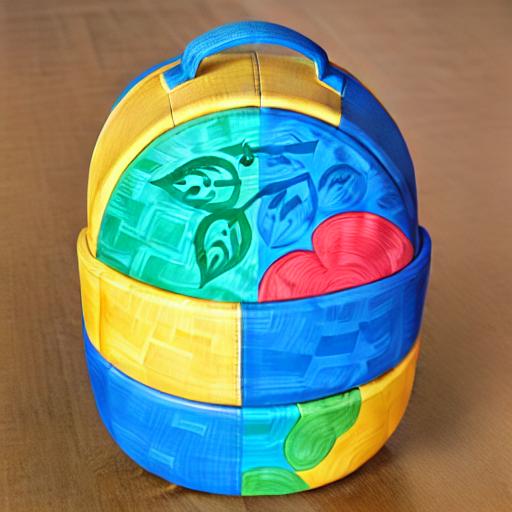} &
        \includegraphics[width=0.117\textwidth,height=0.117\textwidth]{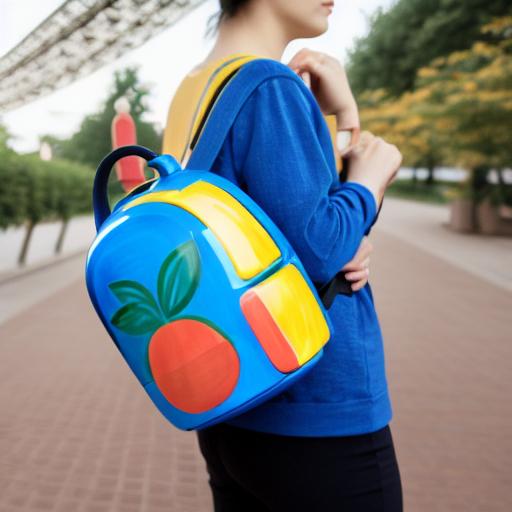} &
        \includegraphics[width=0.117\textwidth,height=0.117\textwidth]{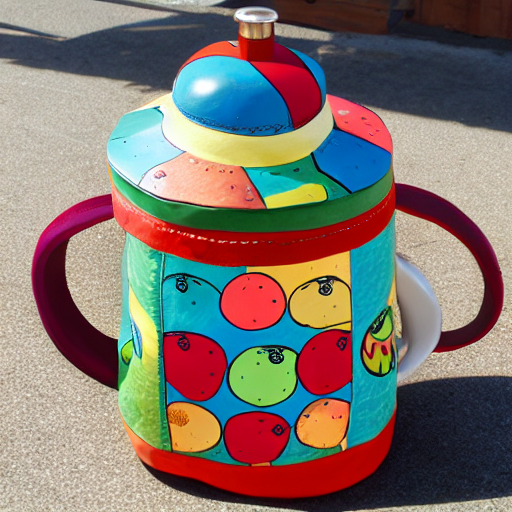} & \includegraphics[width=0.117\textwidth,height=0.117\textwidth]{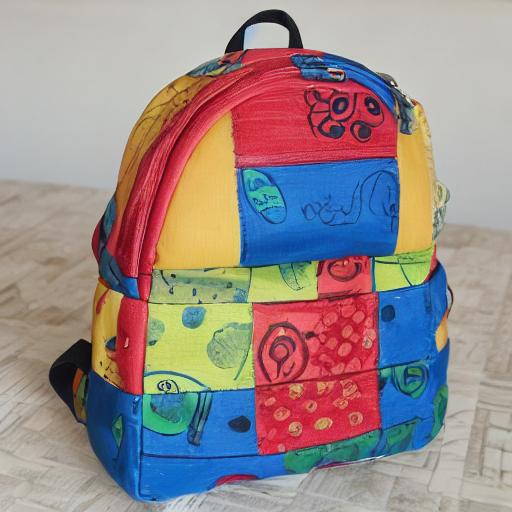} &  
        \\
        
         \includegraphics[width=0.117\textwidth,height=0.117\textwidth]{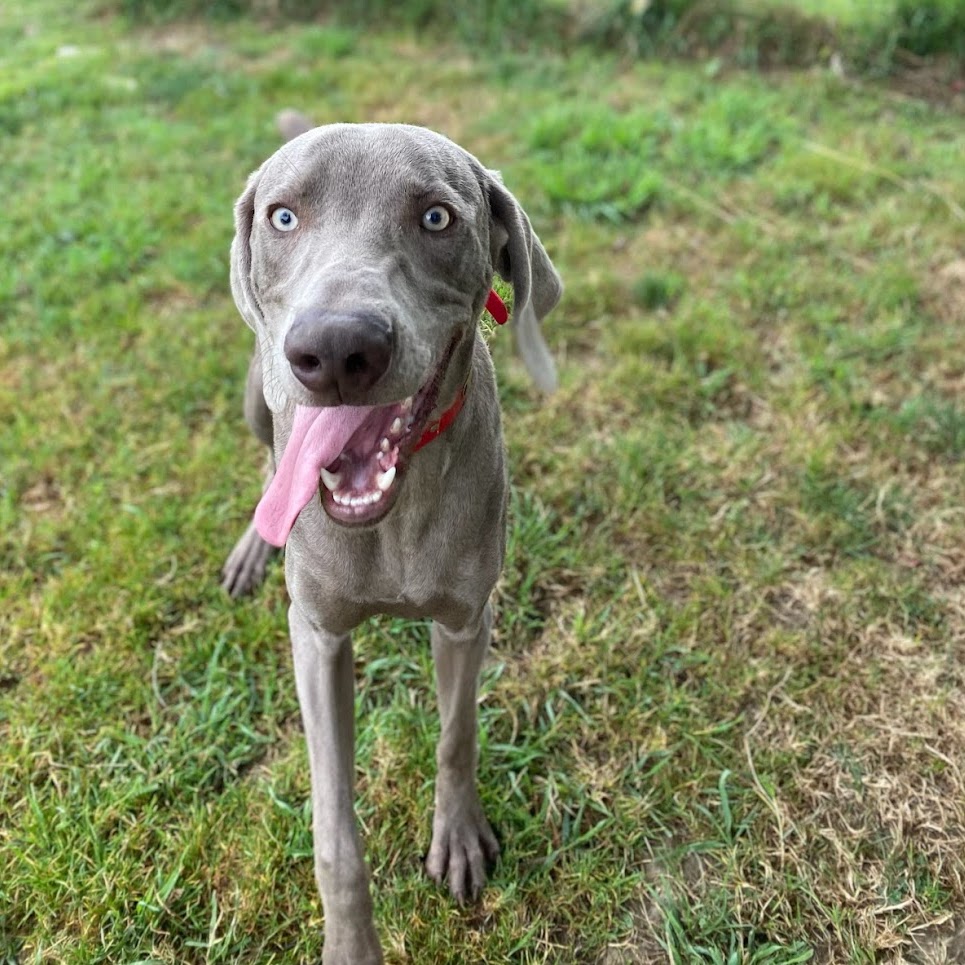} & \raisebox{0.048\textwidth}{\rotatebox[origin=t]{0}{\scalebox{0.9}{\begin{tabular}{c@{}c@{}c@{}} \pholdercolor{} in the style of Monet\end{tabular}}}} &
        \includegraphics[width=0.117\textwidth,height=0.117\textwidth]{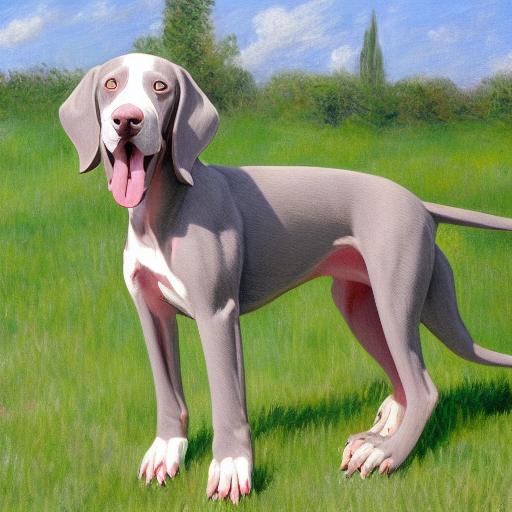} &
        \includegraphics[width=0.117\textwidth,height=0.117\textwidth]{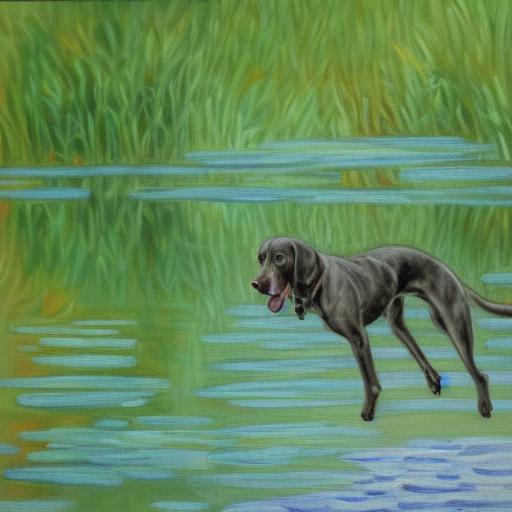} &
        \includegraphics[width=0.117\textwidth,height=0.117\textwidth]{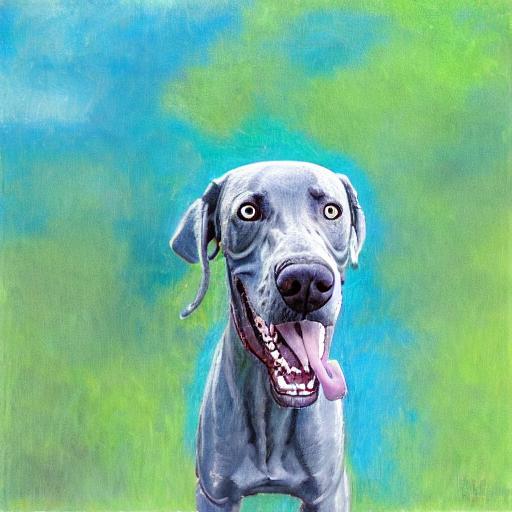} &
        \includegraphics[width=0.117\textwidth,height=0.117\textwidth]{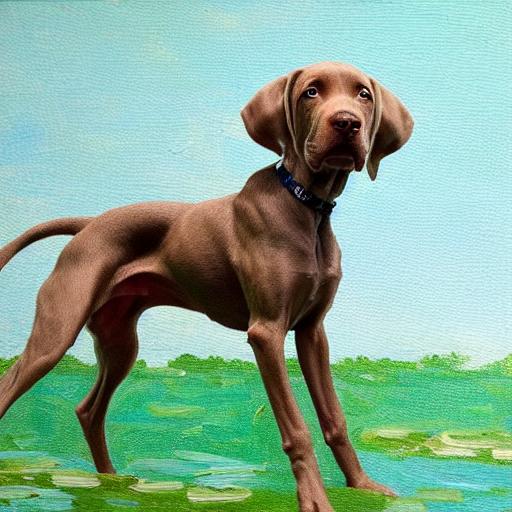} &
        \includegraphics[width=0.117\textwidth,height=0.117\textwidth]{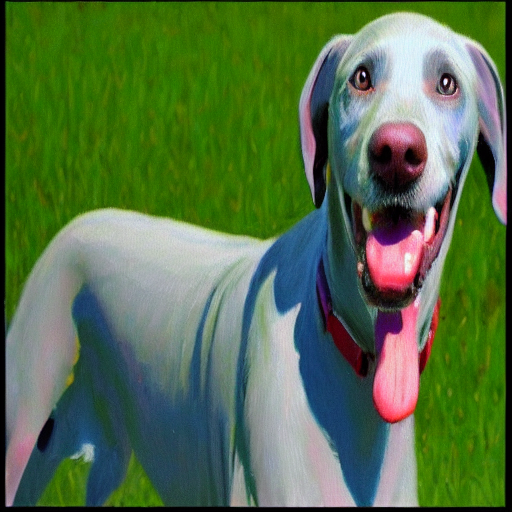} & \includegraphics[width=0.117\textwidth,height=0.117\textwidth]{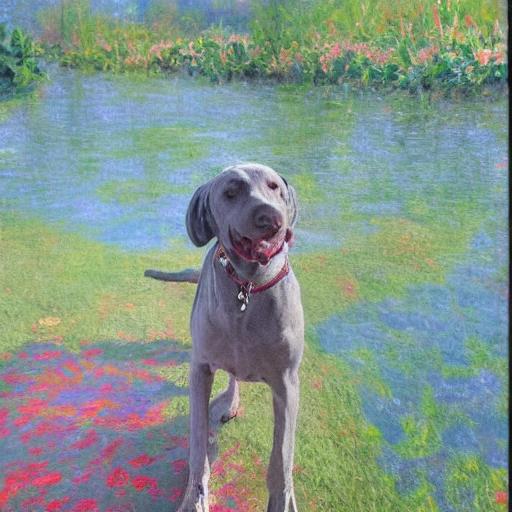} &  
        \\

         \includegraphics[width=0.117\textwidth,height=0.117\textwidth]{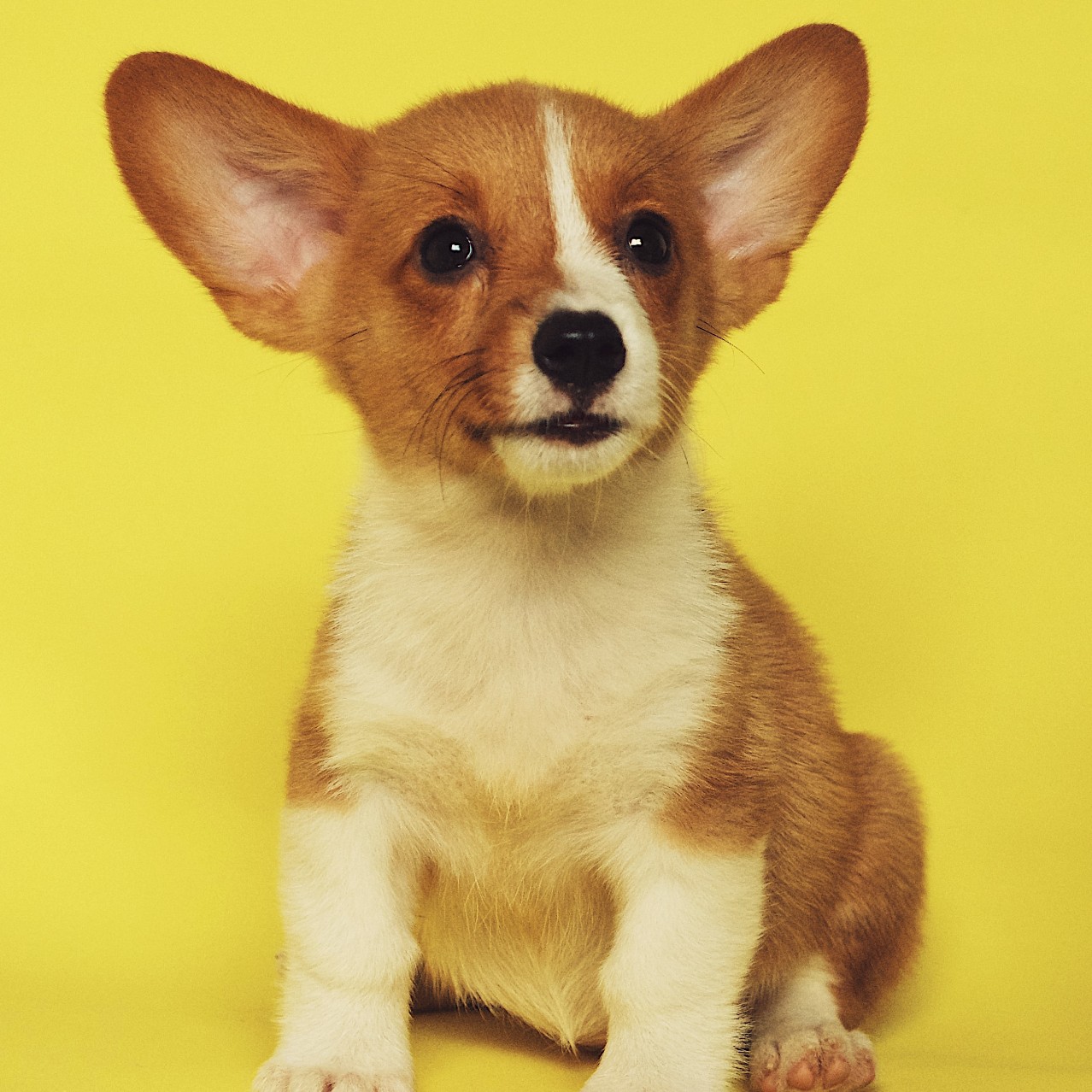} & \raisebox{0.048\textwidth}{\rotatebox[origin=t]{0}{\scalebox{0.9}{\begin{tabular}{c@{}c@{}c@{}} Pixar rendering of \pholdercolor{} \end{tabular}}}} &
        \includegraphics[width=0.117\textwidth,height=0.117\textwidth]{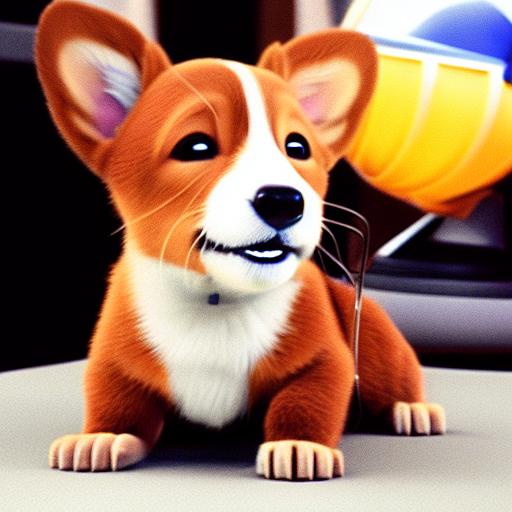} &
        \includegraphics[width=0.117\textwidth,height=0.117\textwidth]{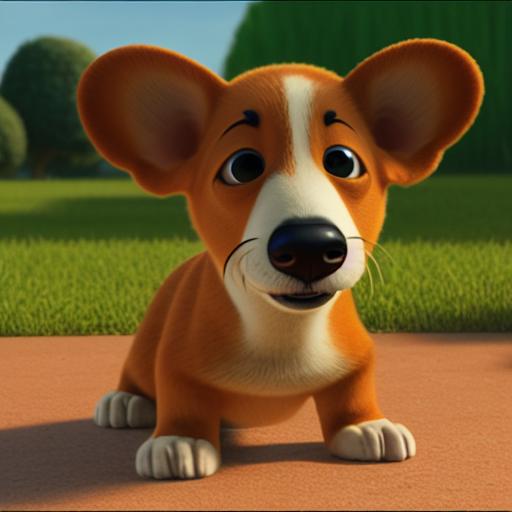} &
        \includegraphics[width=0.117\textwidth,height=0.117\textwidth]{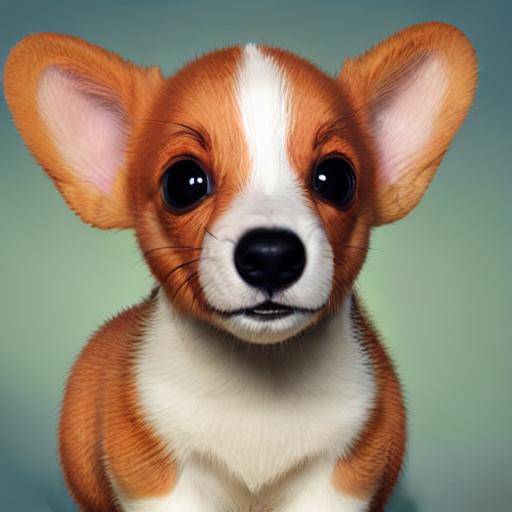} &
        \includegraphics[width=0.117\textwidth,height=0.117\textwidth]{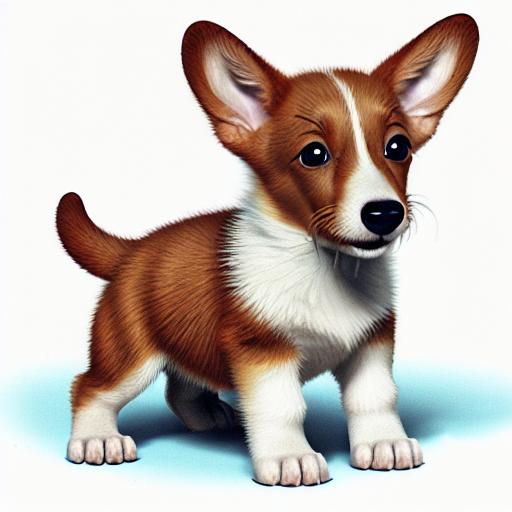} &
        \includegraphics[width=0.117\textwidth,height=0.117\textwidth]{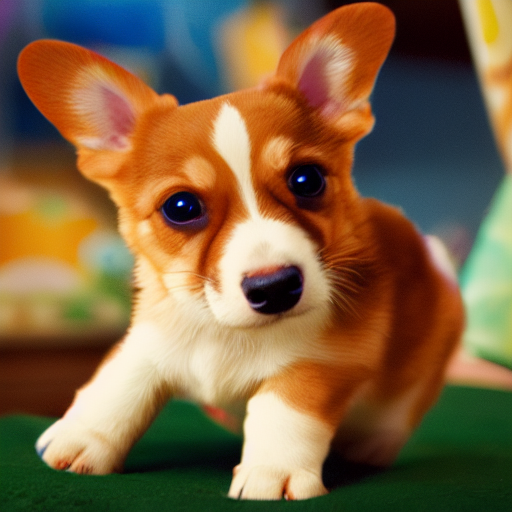} & \includegraphics[width=0.117\textwidth,height=0.117\textwidth]{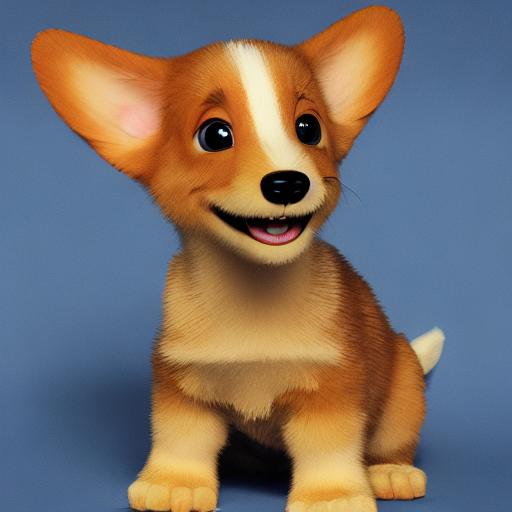} &  
        \\

        \includegraphics[width=0.117\textwidth,height=0.117\textwidth]{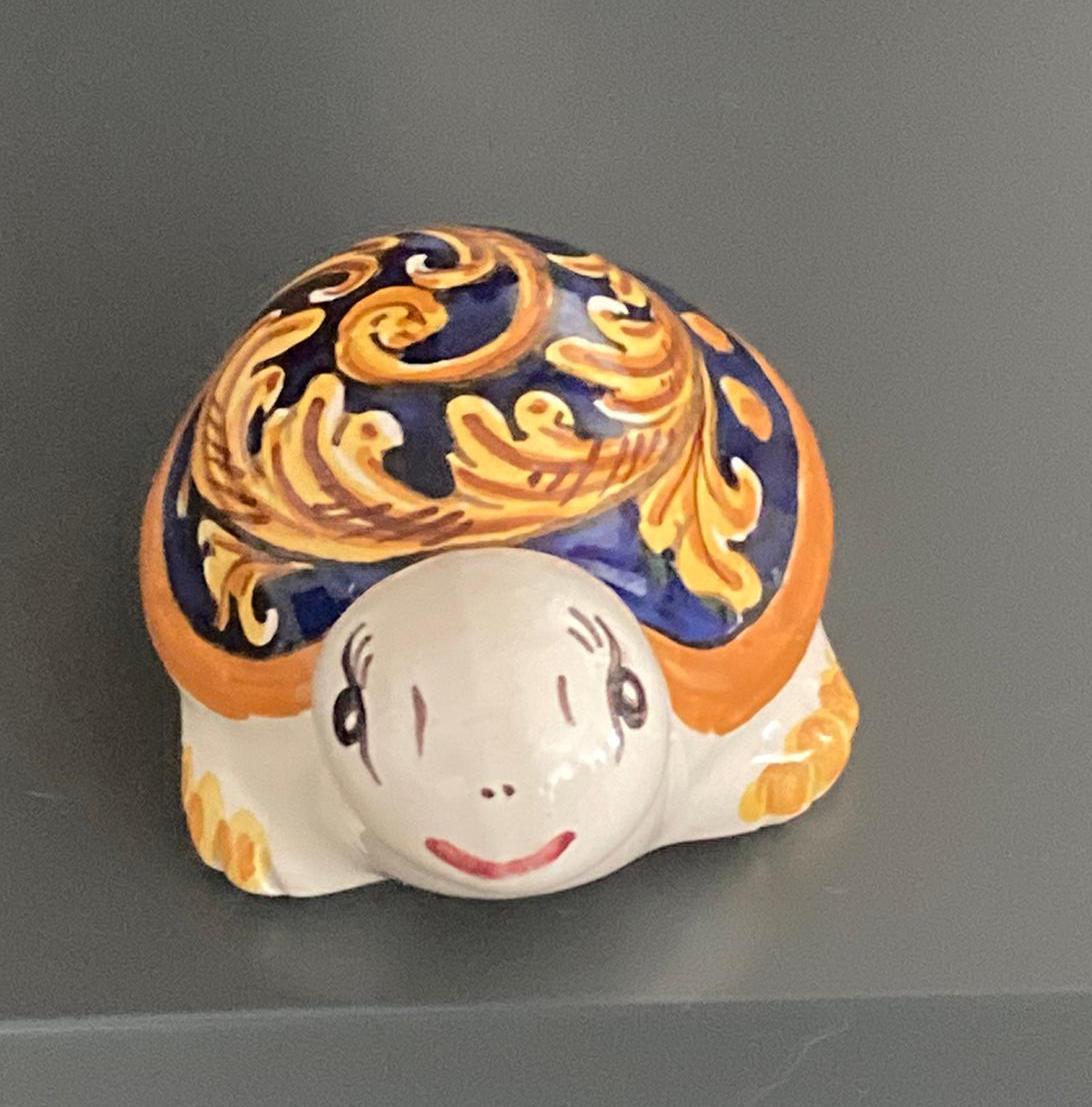} & \raisebox{0.048\textwidth}{\rotatebox[origin=t]{0}{\scalebox{0.9}{\begin{tabular}{c@{}c@{}c@{}} Vector art of \pholdercolor{}\end{tabular}}}} &
        \includegraphics[width=0.117\textwidth,height=0.117\textwidth]{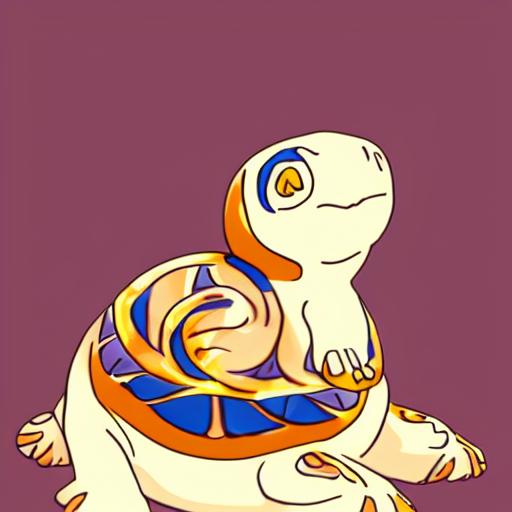} &
        \includegraphics[width=0.117\textwidth,height=0.117\textwidth]{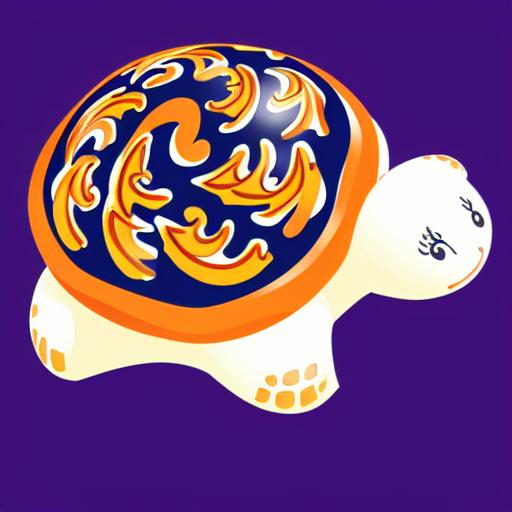} &
        \includegraphics[width=0.117\textwidth,height=0.117\textwidth]{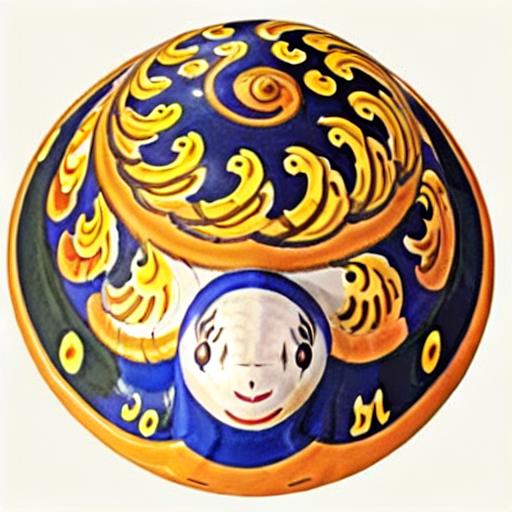} &
        \includegraphics[width=0.117\textwidth,height=0.117\textwidth]{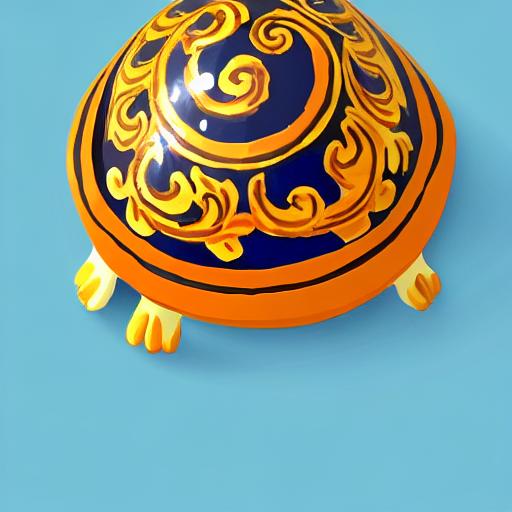} &
        \includegraphics[width=0.117\textwidth,height=0.117\textwidth]{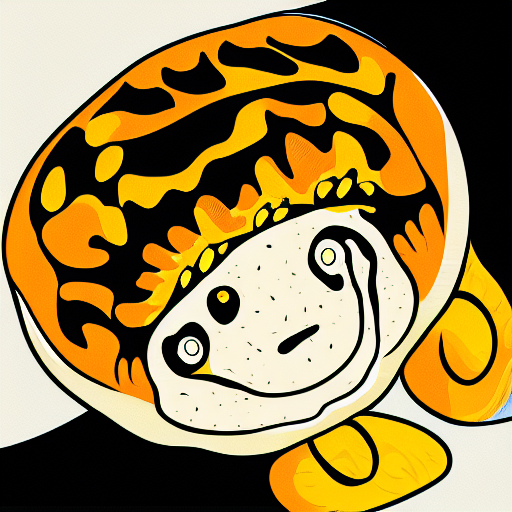} & \includegraphics[width=0.117\textwidth,height=0.117\textwidth]{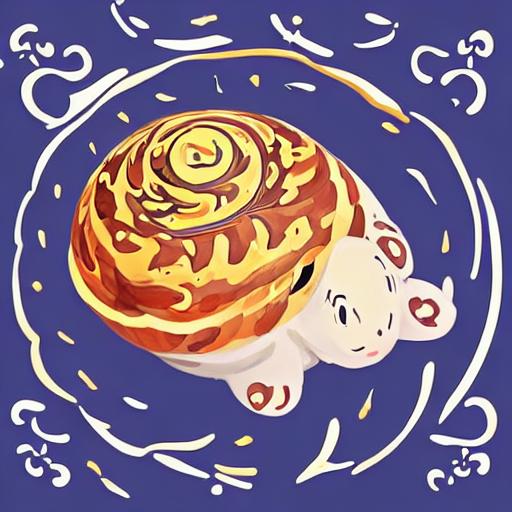} &  
        \\

        \includegraphics[width=0.117\textwidth,height=0.117\textwidth]{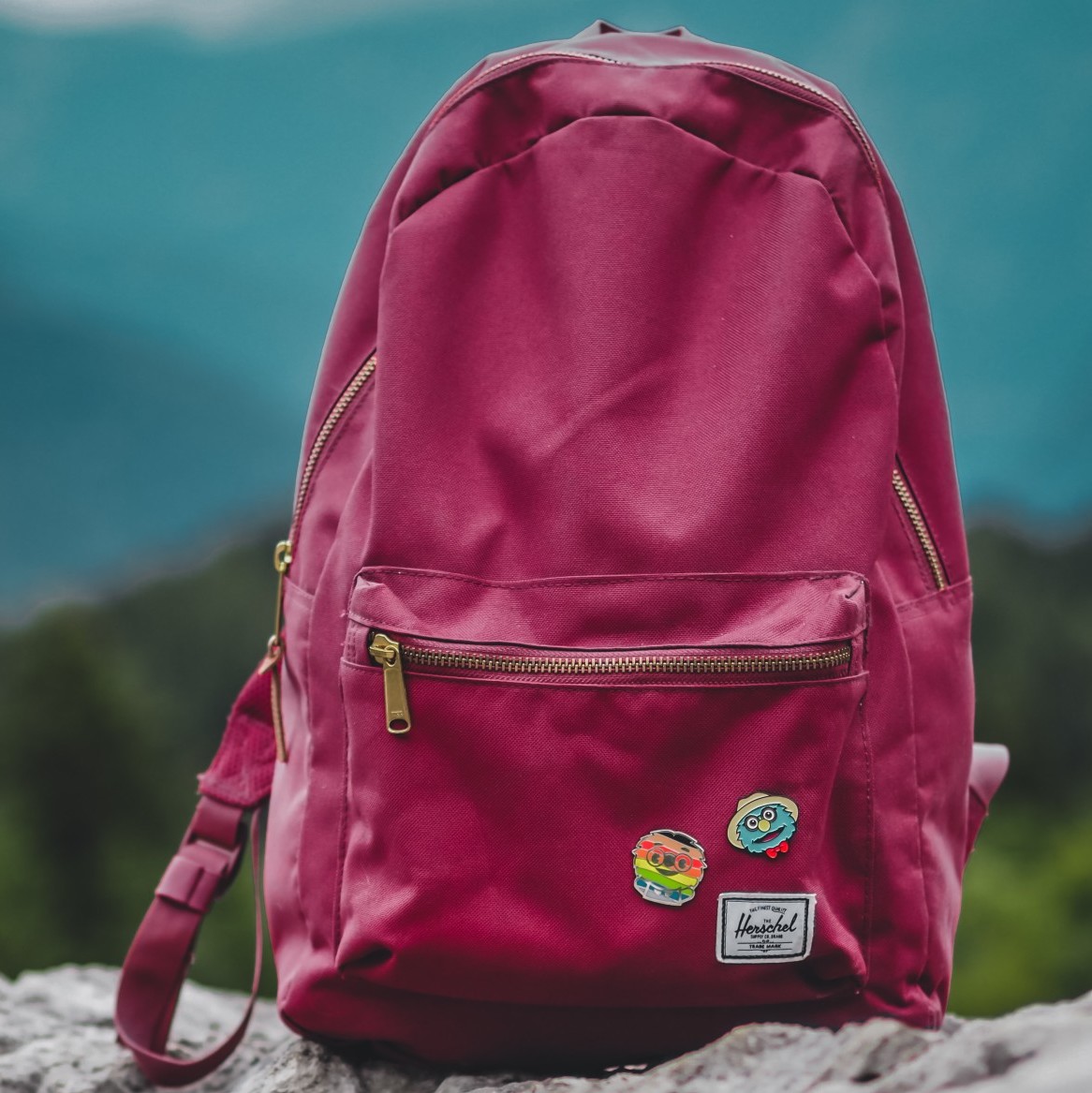} & \raisebox{0.048\textwidth}{\rotatebox[origin=t]{0}{\scalebox{0.9}{\begin{tabular}{c@{}c@{}c@{}} \pholdercolor{} near Mount Fuji \end{tabular}}}} &
        \includegraphics[width=0.117\textwidth,height=0.117\textwidth]{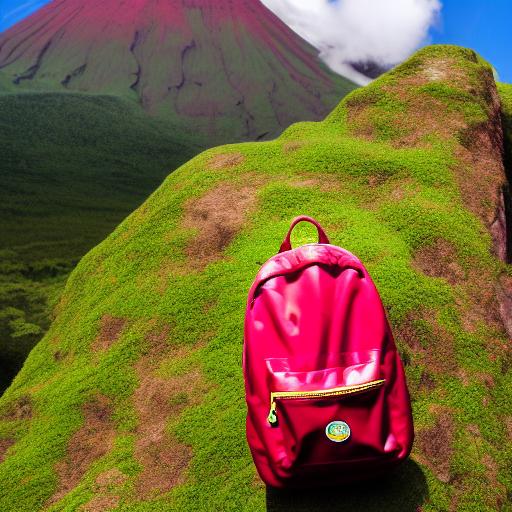} &
        \includegraphics[width=0.117\textwidth,height=0.117\textwidth]{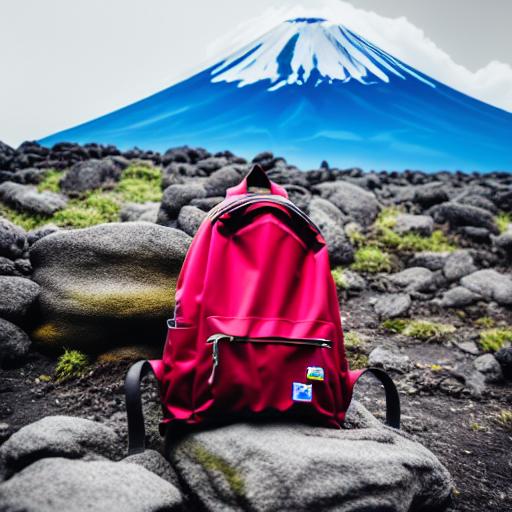} &
        \includegraphics[width=0.117\textwidth,height=0.117\textwidth]{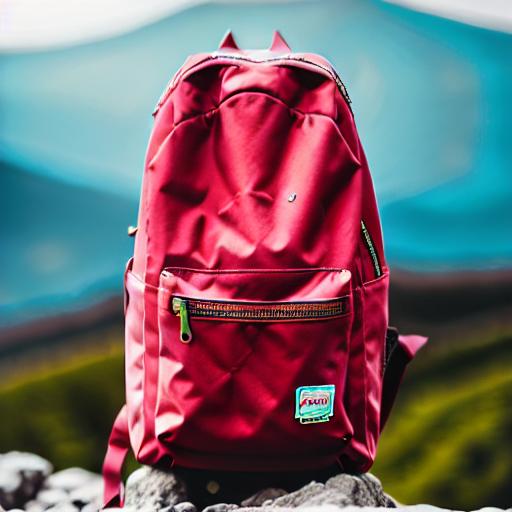} &
        \includegraphics[width=0.117\textwidth,height=0.117\textwidth]{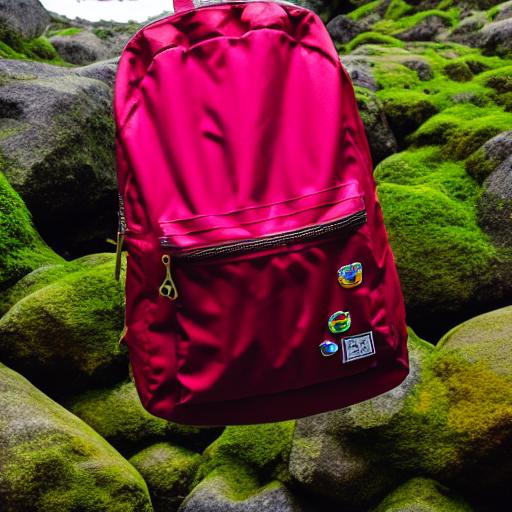} &
        \includegraphics[width=0.117\textwidth,height=0.117\textwidth]{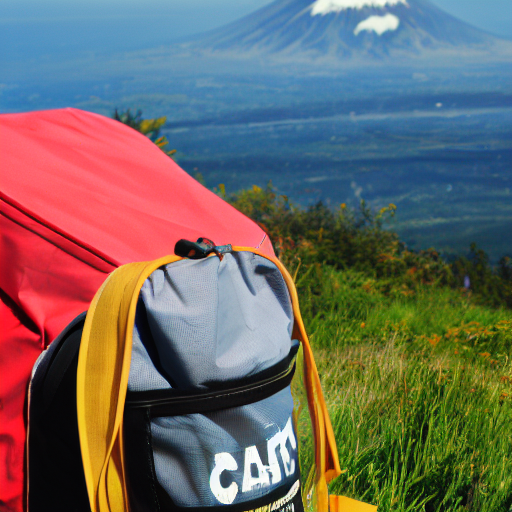} & \includegraphics[width=0.117\textwidth,height=0.117\textwidth]{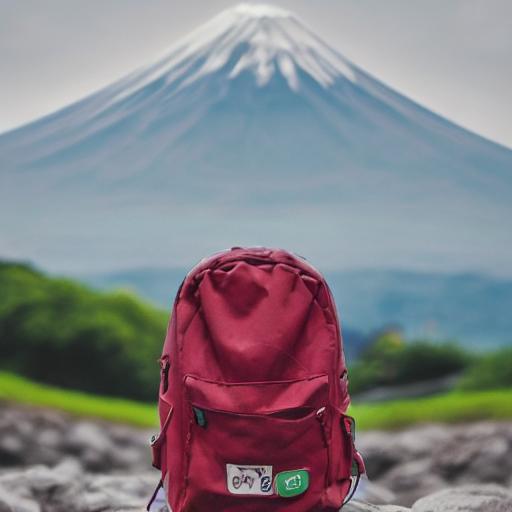} &  
        \\

       \includegraphics[width=0.117\textwidth,height=0.117\textwidth]{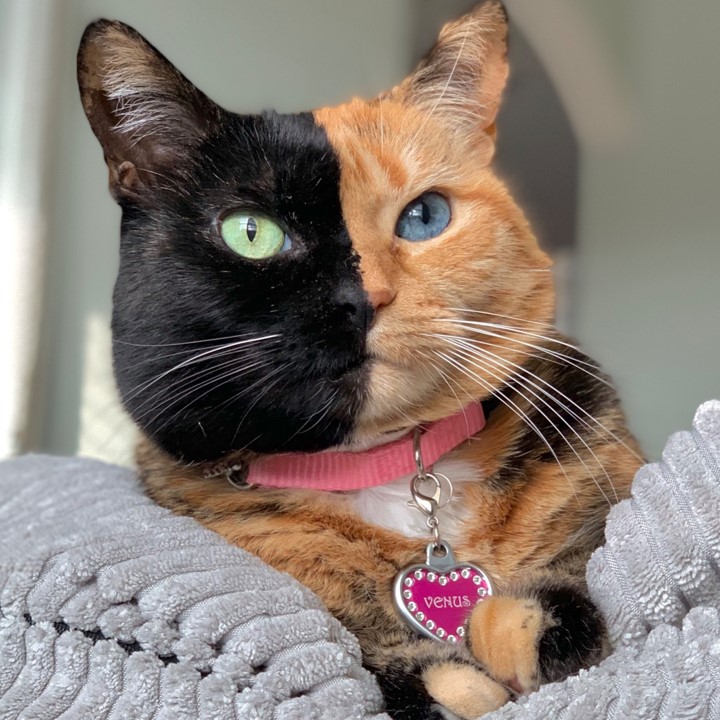} & \raisebox{0.048\textwidth}{\rotatebox[origin=t]{0}{\scalebox{0.9}{\begin{tabular}{c@{}c@{}c@{}} Ukiyo-e painting of \pholdercolor{} \end{tabular}}}} &
        \includegraphics[width=0.117\textwidth,height=0.117\textwidth]{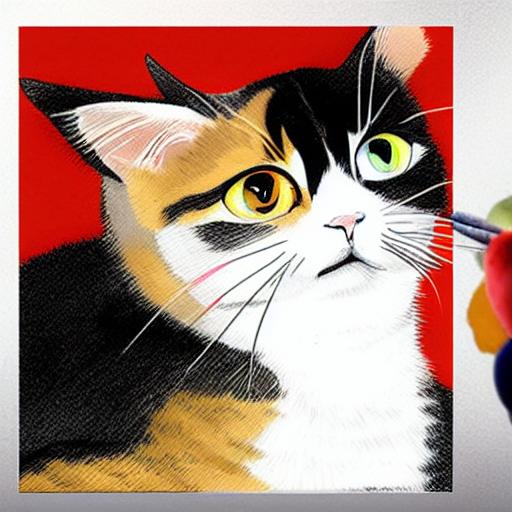} &
        \includegraphics[width=0.117\textwidth,height=0.117\textwidth]{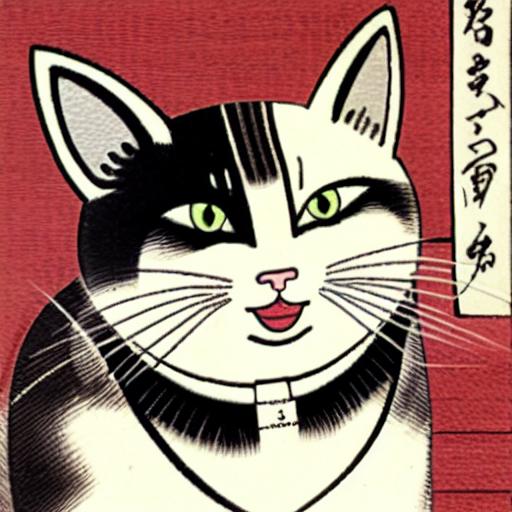} &
        \includegraphics[width=0.117\textwidth,height=0.117\textwidth]{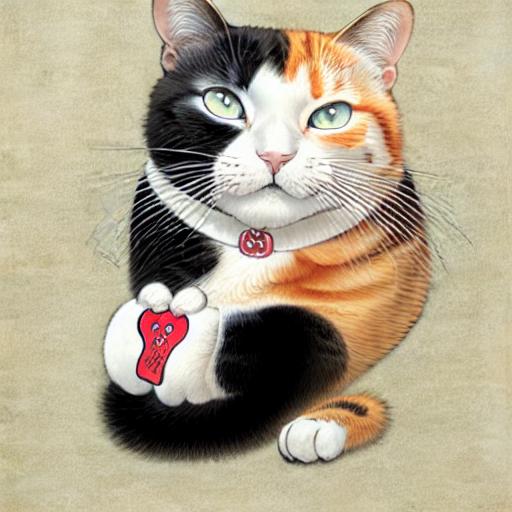} &
        \includegraphics[width=0.117\textwidth,height=0.117\textwidth]{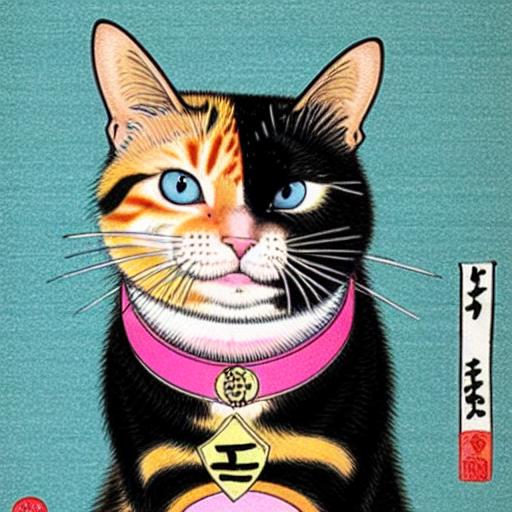} &
        \includegraphics[width=0.117\textwidth,height=0.117\textwidth]{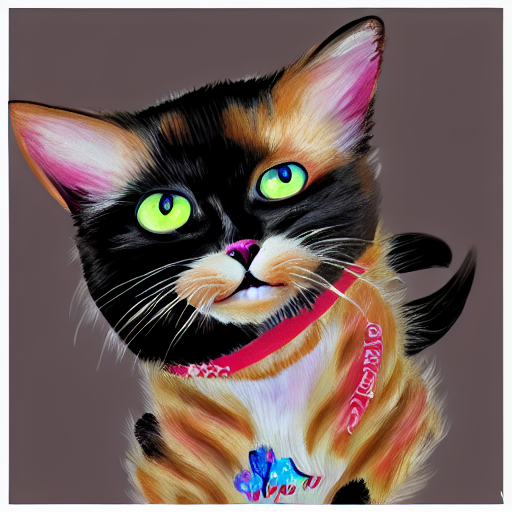} & \includegraphics[width=0.117\textwidth,height=0.117\textwidth]{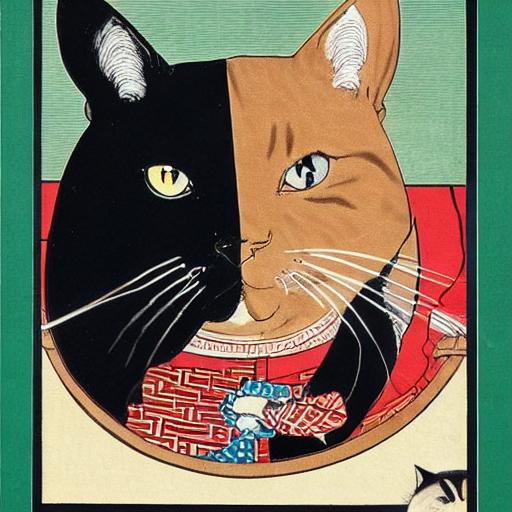} &  
        \\

        \includegraphics[width=0.117\textwidth,height=0.117\textwidth]{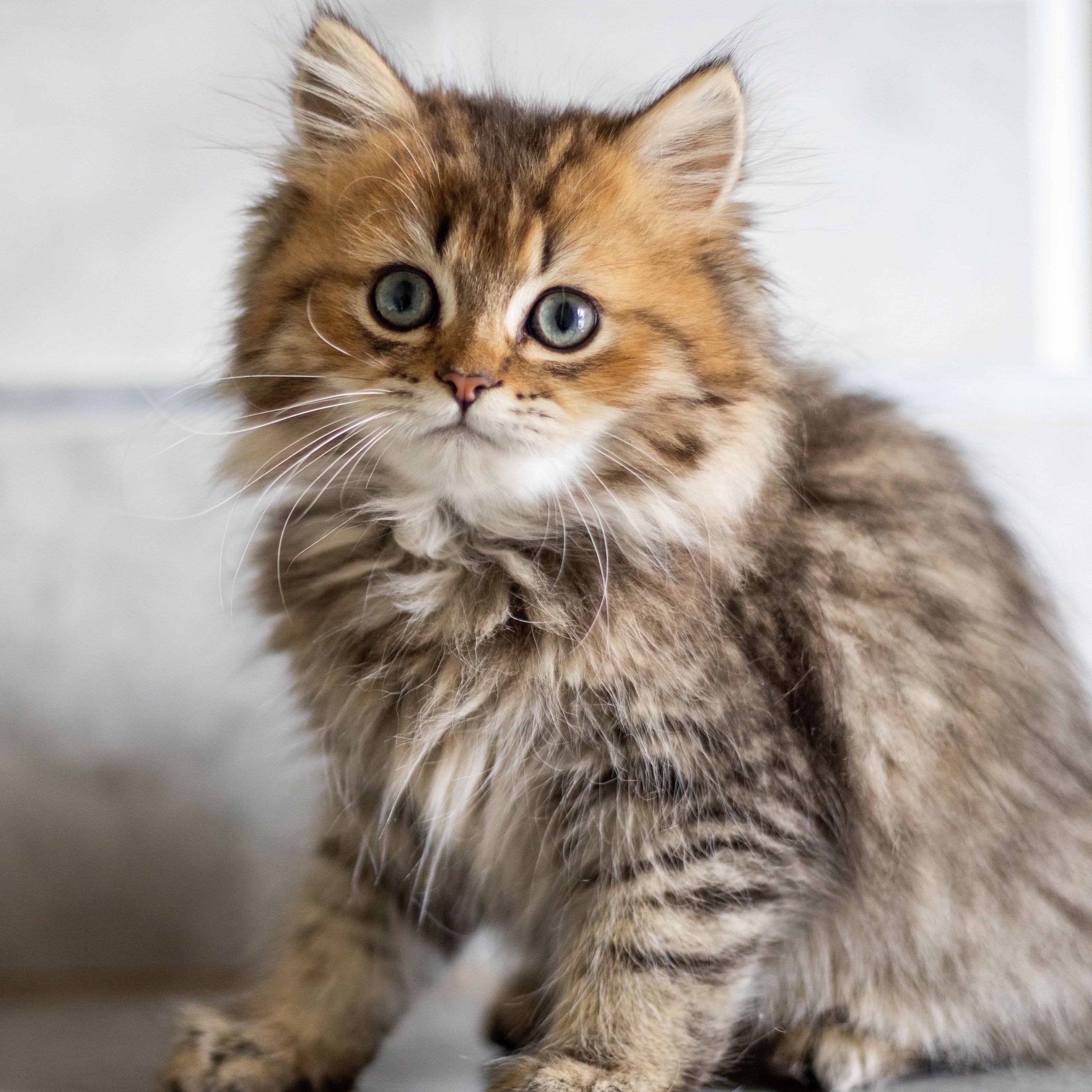} & \raisebox{0.048\textwidth}{\rotatebox[origin=t]{0}{\scalebox{0.9}{\begin{tabular}{c@{}c@{}c@{}} \pholdercolor{} with a tophat\end{tabular}}}} &
        \includegraphics[width=0.117\textwidth,height=0.117\textwidth]{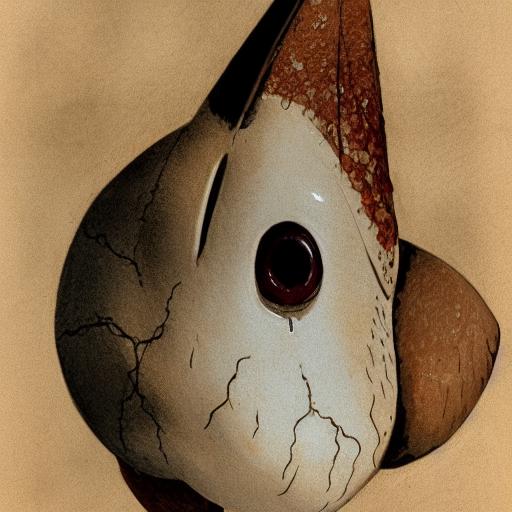} &
        \includegraphics[width=0.117\textwidth,height=0.117\textwidth]{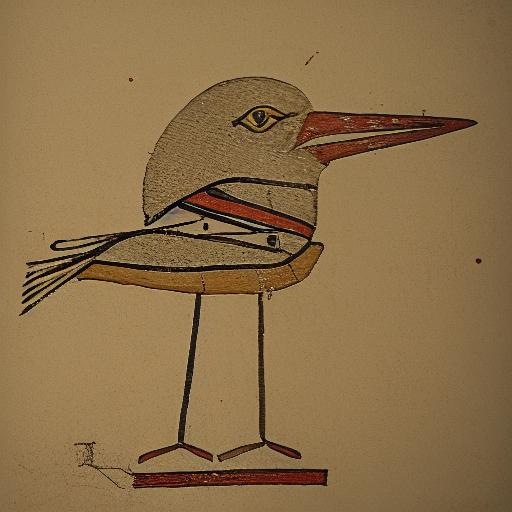} &
        \includegraphics[width=0.117\textwidth,height=0.117\textwidth]{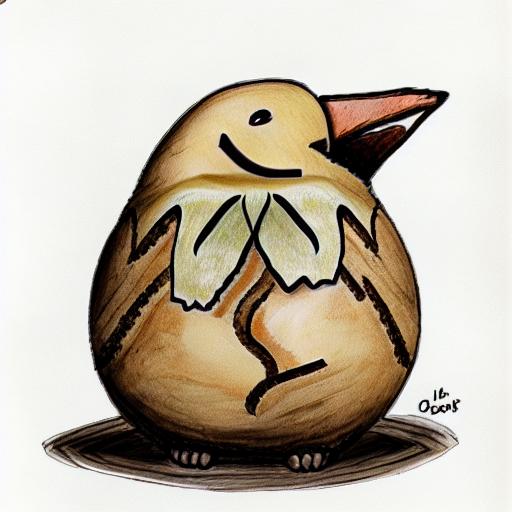} &
        \includegraphics[width=0.117\textwidth,height=0.117\textwidth]{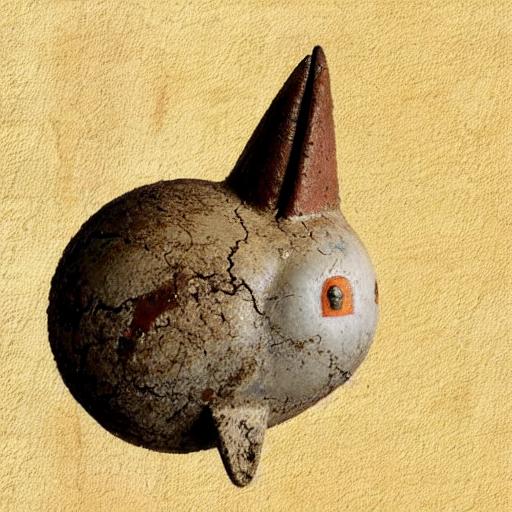} &
        \includegraphics[width=0.117\textwidth,height=0.117\textwidth]{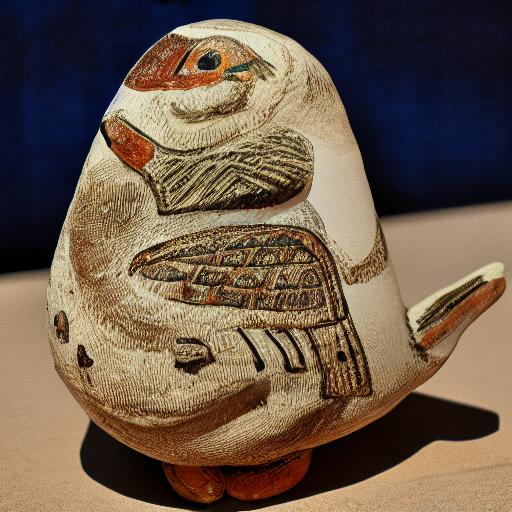} & \includegraphics[width=0.117\textwidth,height=0.117\textwidth]{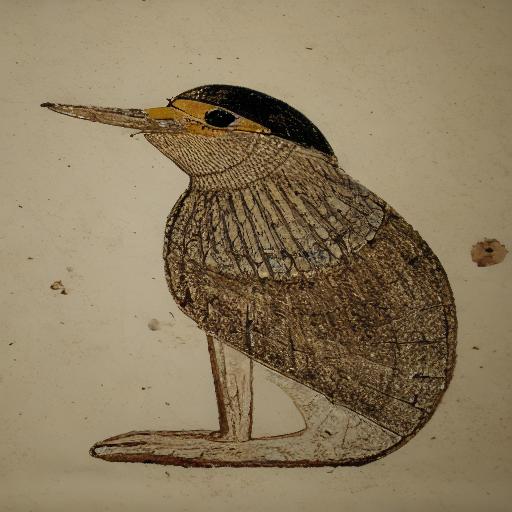} &  
        \\

        \includegraphics[width=0.117\textwidth,height=0.117\textwidth]{figures/comparison/fluffy-cat/input.jpg} & \raisebox{0.048\textwidth}{\rotatebox[origin=t]{0}{\scalebox{0.9}{\begin{tabular}{c@{}c@{}c@{}} \pholdercolor{} with a tophat\end{tabular}}}} &
        \includegraphics[width=0.117\textwidth,height=0.117\textwidth]{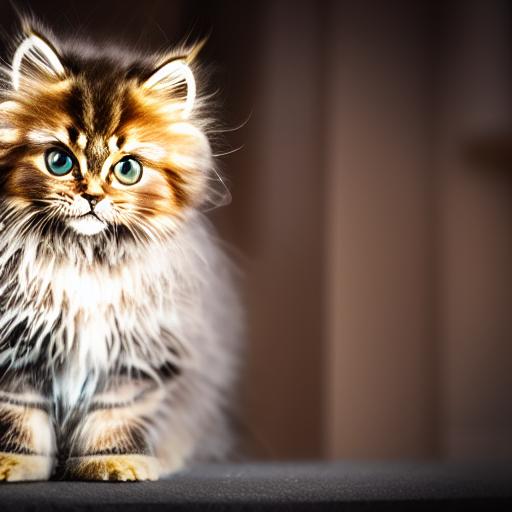} &
        \includegraphics[width=0.117\textwidth,height=0.117\textwidth]{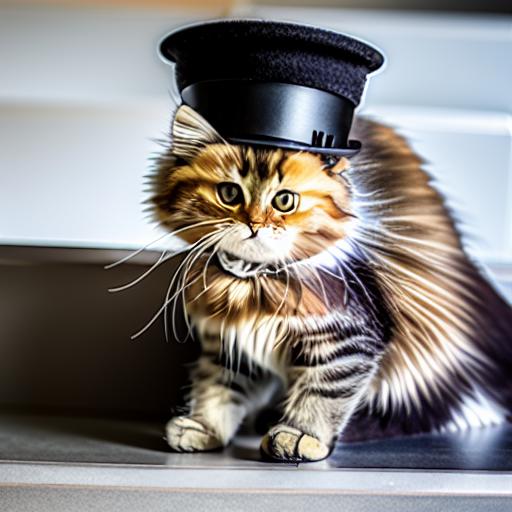} &
        \includegraphics[width=0.117\textwidth,height=0.117\textwidth]{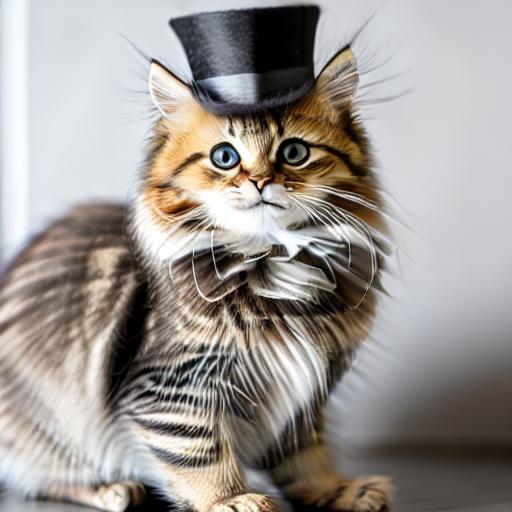} &
        \includegraphics[width=0.117\textwidth,height=0.117\textwidth]{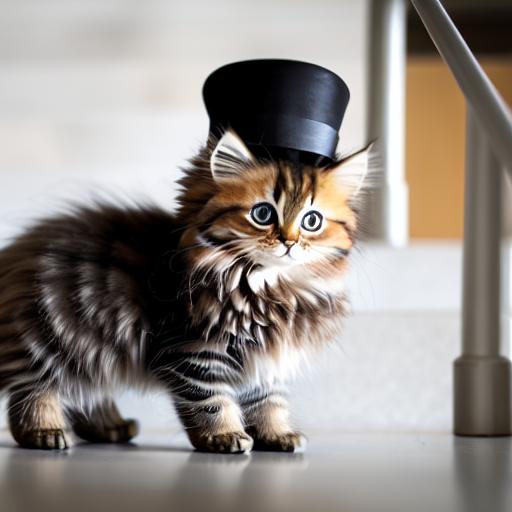} &
        \includegraphics[width=0.117\textwidth,height=0.117\textwidth]{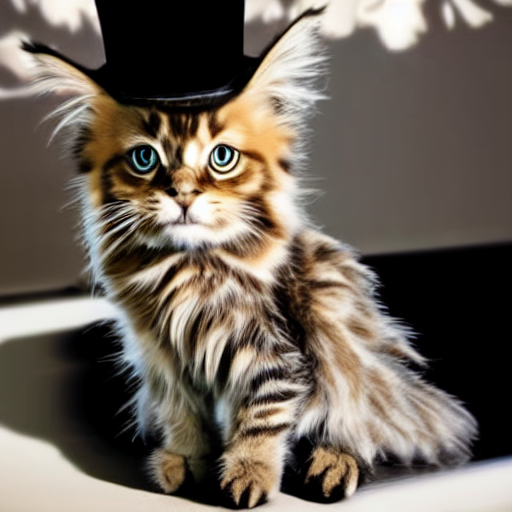} & \includegraphics[width=0.117\textwidth,height=0.117\textwidth]{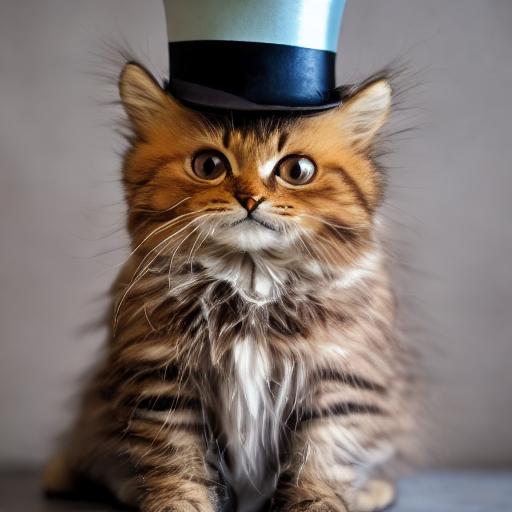} &  
        \\

        \includegraphics[width=0.117\textwidth,height=0.117\textwidth]{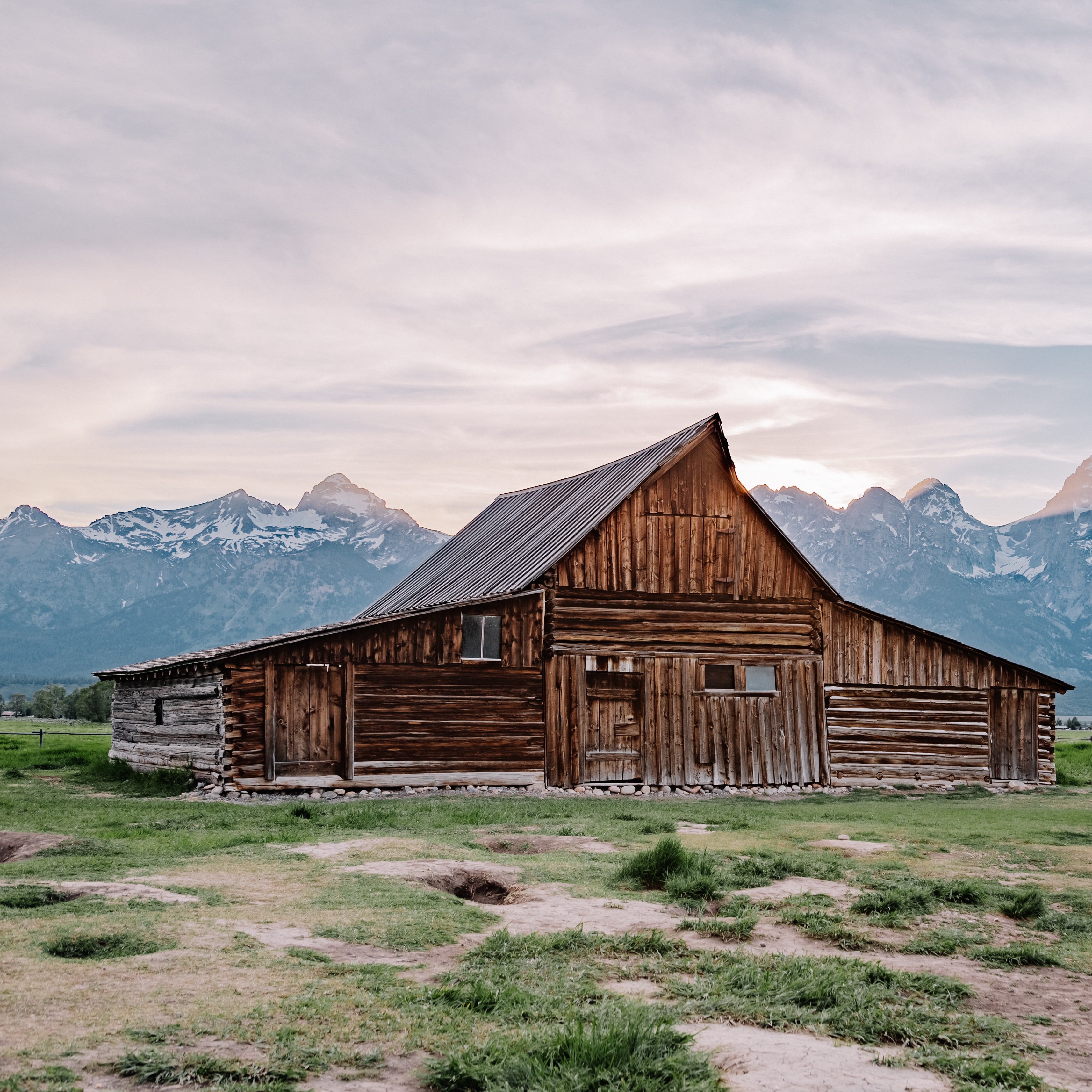} & \raisebox{0.048\textwidth}{\rotatebox[origin=t]{0}{\scalebox{0.9}{\begin{tabular}{c@{}c@{}c@{}} Manga drawing of \pholdercolor{} \end{tabular}}}} &
        \includegraphics[width=0.117\textwidth,height=0.117\textwidth]{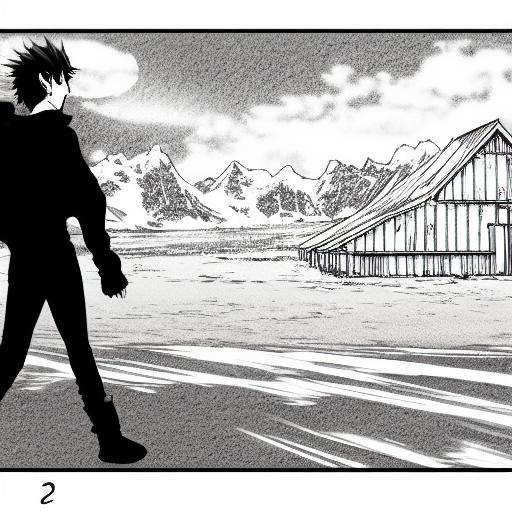} &
        \includegraphics[width=0.117\textwidth,height=0.117\textwidth]{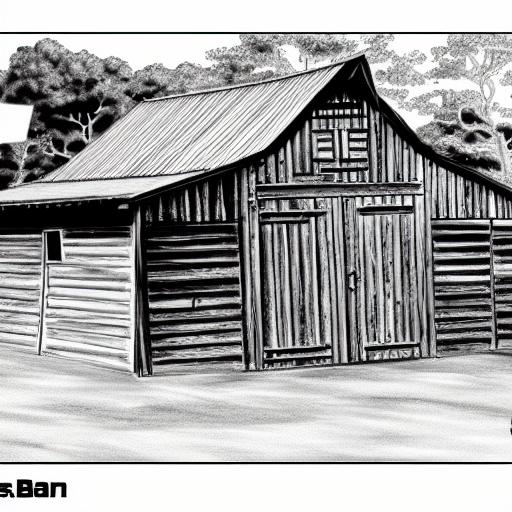} &
        \includegraphics[width=0.117\textwidth,height=0.117\textwidth]{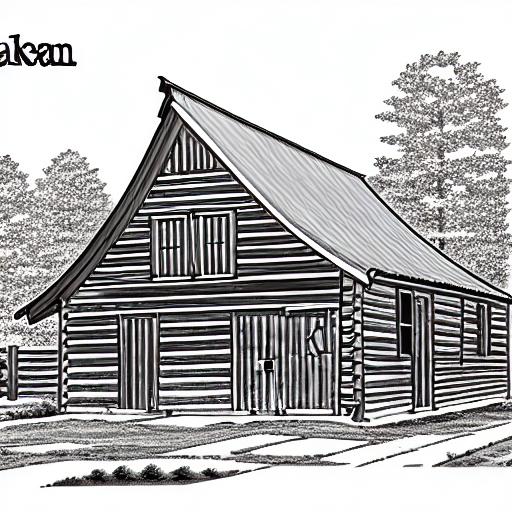} &
        \includegraphics[width=0.117\textwidth,height=0.117\textwidth]{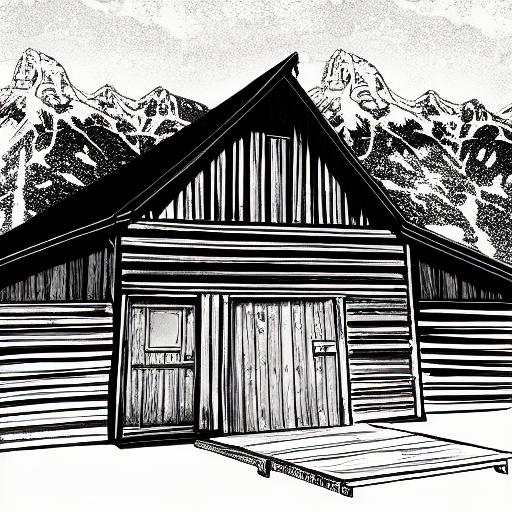} &
        \includegraphics[width=0.117\textwidth,height=0.117\textwidth]{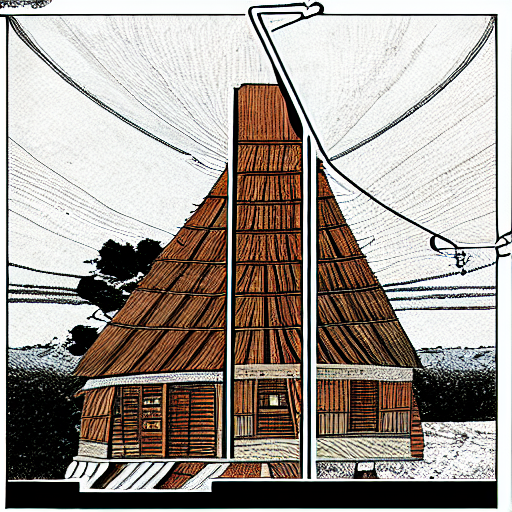} & \includegraphics[width=0.117\textwidth,height=0.117\textwidth]{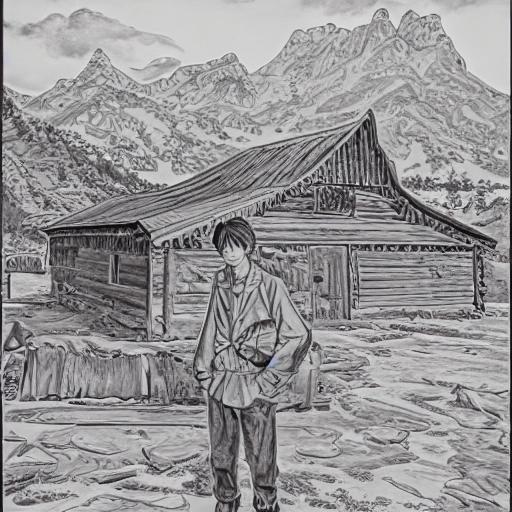} &  
        \\

    \end{tabular}
    
    }
    \caption{Qualitative comparison with existing methods. Our method achieves comparable quality to the state-of-the-art using only a single image and $12$ or fewer training steps. Notably, it generalizes to unique objects which recent encoder-based methods struggle with.}
    \label{fig:qual_comp}
\end{figure*}
\FloatBarrier
\subsection{Hyper-weights Regularization}

Our encoder also contains a hypernetwork branch, whose predicted weights can also overfit the model to a given image~\citep{gal2023designing}. To address the issue of overfitting caused by the hyper-network predictions, we propose a modification to the UNET forward pass. We begin by duplicating each block into two copies. The first block uses the original UNET's weights, and for the second, we use the hypernetwork-modulated weights. Moreover, in the first (original weight) branch, we replace our predicted word embeddings with those of the nearest neighbor token. The outputs of the two paths are then linearly blended with a coefficient of $\alpha_{blend}$. This dual-call approach ensures that one path is free from attention-overfitting, and can thereby strike a balance between capturing the identity and preserving the model's prior knowledge (see Fig~\ref{fig:arch-fig}).

Specifically, given the weight modulations $W_{\Delta}$ and the predicted word embedding $v_*$ from our encoder $E$, we first identify the nearest hard-token embedding $v_h$ to the model's prediction $v_*$. We then compose two text prompts, $C$ and $C_h$, which consist of $v_*$ and $v_h$ respectively. In other words, $C$ and  $C_h$ are derived from the same prompt, but one uses the learned embedding while the other uses only real-token ("hard") embeddings.
 
For each block $B$ of the UNET-denoiser, which receives a feature map $f\in \mathbb{R}^{k\times k \times D}$, text condition $C$, and weight modulation $W_{\Delta}$, we modify the block using the dual-path approach:
\begin{equation}
\label{eq:dual_adaptation}
    out = \alpha_{blend} \cdot B(f,C,W_\Delta)+  (1-\alpha_{blend}) \cdot B(f,C_{h},\emptyset)
\end{equation}.

\subsection{Inference-time Personalization}
As a final step, we follow E4T and employ a brief tuning phase at inference time. While E4T tunes both the model and the encoder at inference time, we find that this proccess requires significant memory (roughly 70GB with the recommended minimal batch size of $16$). To reduce this requirement, we note that our model predicts the same embedding and weight decomposition used by LoRA~\cite{simoLoRA2023, Hu2021LoRALA}. As such, we can use its output as an initialization for a short LoRA-tuning run, with the addition of an L2-regularization term that aims to keep both weights and embedding close to the original encoder prediction.
\section{Experiments}

\subsection{Experimental setup}
\paragraph{\textbf{Pre-training:}} We initiated our experiments by pre-training our model on the ImageNet-1K and Open-Images datasets~\cite{ILSVRC15,  OpenImages}. ImageNet-1K consists of 1.28 million training images from 1,000 distinct classes. For OpenImages dataset, we crop the largest object from each training image to avoid training on multiple-subjects at the same time. Together, our training data consists of around 3M images. The pre-training phase employed a pre-trained CLIP model with a ViT-H-14 encoder as the backbone architecture. The token-embedder and hyper-network were trained using a learning rate of $\text{lr=1e-4}$ with linear warm-up and cosine-decay scheduling. For ablation purposes, we conducted 50,000 iterations during training. For our final model and comparisons to prior art, we extended the training to 150,000 steps.

\paragraph{\textbf{Inference-tuning Phase:}} During the inference-time tuning phase, we used a single-forward pass to obtain the initial prediction of the hyper-weights and word-embedding for the text-to-image model adaptation. Subsequently, we optimized the initial prediction using a learning rate of $\text{lr=2e-3}$ and a balancing factor of $\alpha_{blend} = 0.25$ (see Eq.~\ref{eq:dual_adaptation}). We found that up to 12 optimization steps were sufficient to achieve satisfactory results for various concepts, compared to the recommended $2,000$ for LoRA-PTI~\citep{simoLoRA2023,roich2021pivotal}.

\paragraph{\textbf{Evaluation Metric:}} We follow TI~\cite{gal2022image} and employ a CLIP text-to-image similarity score as the evaluation metric to assess the proximity of the generated images to the input prompt. To measure identity preservation, we utilized the image-to-image CLIP similarity loss between the single-image training set and the generated results. All reported metrics are based on a pre-trained ViT-B-16 model. Our evaluation set contains 17 images taken from prior work~\citep{gal2022image,ruiz2022dreambooth,kumari2022multi}. These cover diverse categories ranging from pets (e.g., dogs) to personal items (e.g., backpacks) and even buildings. 

\subsection{The importance of contrastive regularization} 
\begin{figure}[!htb]
    \centering
    \includegraphics[width=\linewidth]{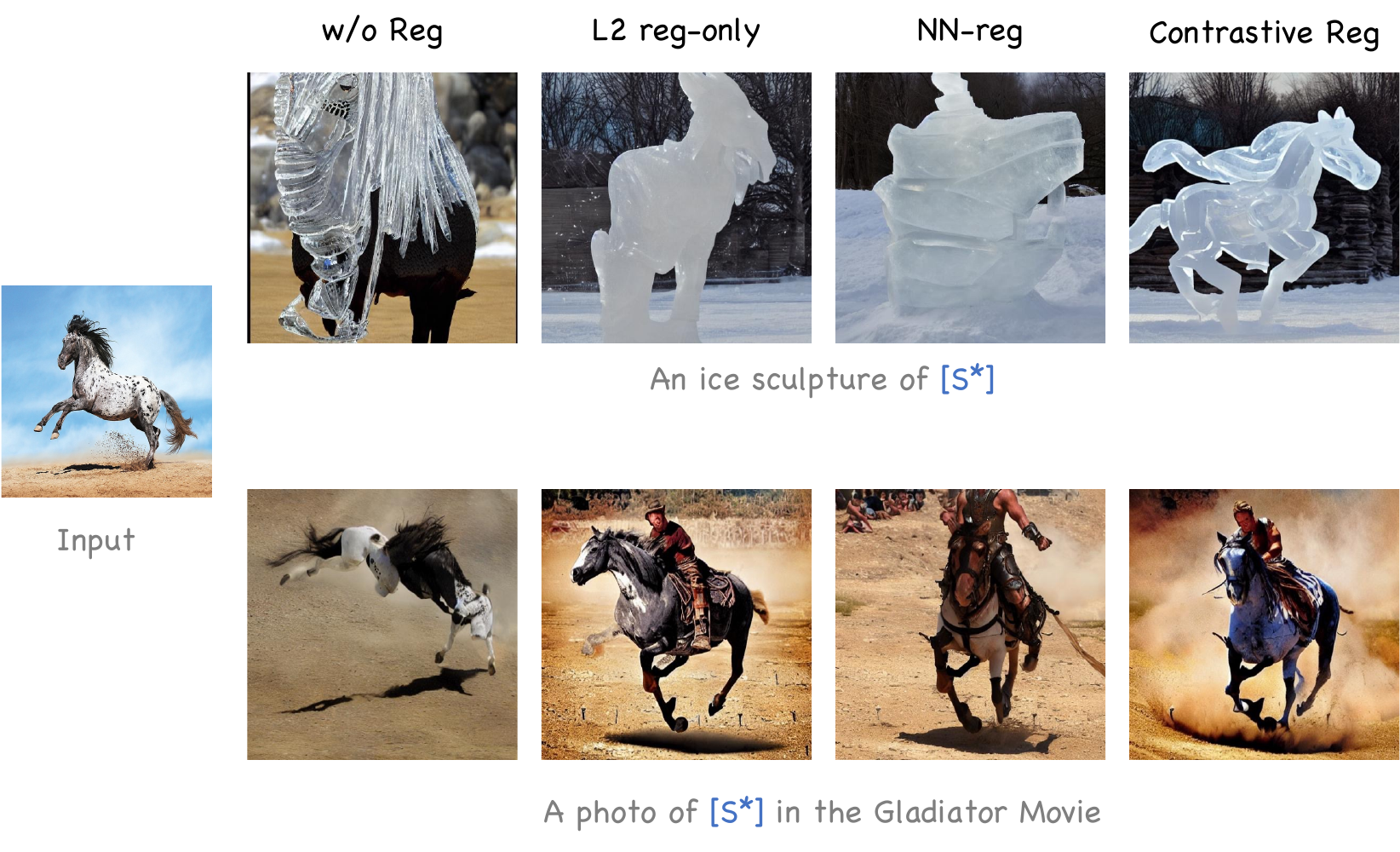}
    \caption{The effects of removing or changing the embedding regularization. Removal of regularization leads to overfitting or mode collapse with poor quality results. \Naive regularizations tend to struggle with preserving the concept details. Our contrastive-based regularization can achieve a tradeoff between the two.}
    \label{fig:contrastive_qualitative}
\end{figure}
Our approach utilizes contrastive learning to improve the quality of predicted embeddings. To visualize the benefit of this regularization, we train our model in four settings: First, without any regularization. Second, we omit all regularization except for the L2 loss on the predicted embedding. Third, we replace the contrastive loss with one that minimizes the cosine-distance between predicted embeddings and their nearest neighbor - a loss inspired by the codebook losses employed in VQGAN~\citep{VQGANEsser}. Finally, we use our proposed contrastive-based alternative.

\begin{figure*}[!hbt]
     \centering
     \begin{subfigure}[b]{0.48\textwidth}
         \centering
         \includegraphics[width=\textwidth]{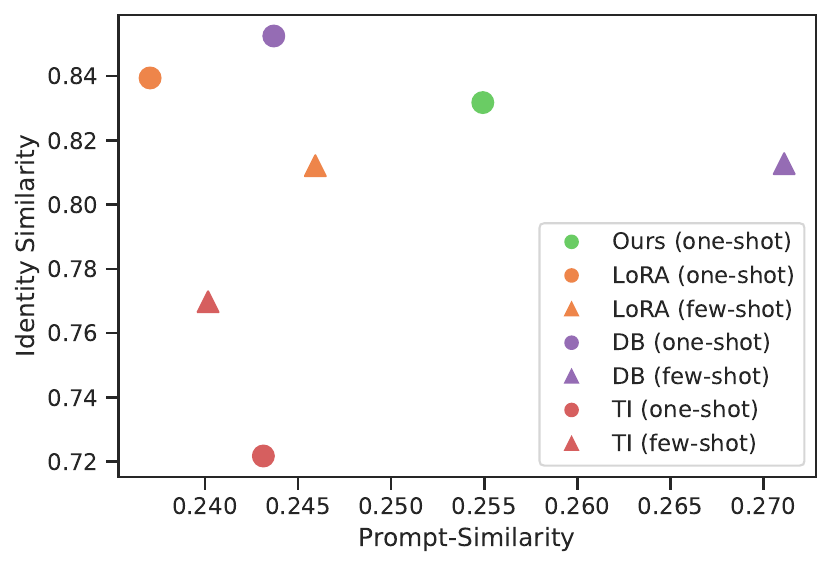}
         \caption{}
         \label{fig:comparison_quant}
     \end{subfigure}
     \hfill
     \begin{subfigure}[b]{0.48\textwidth}
         \centering
         \includegraphics[width=\textwidth]{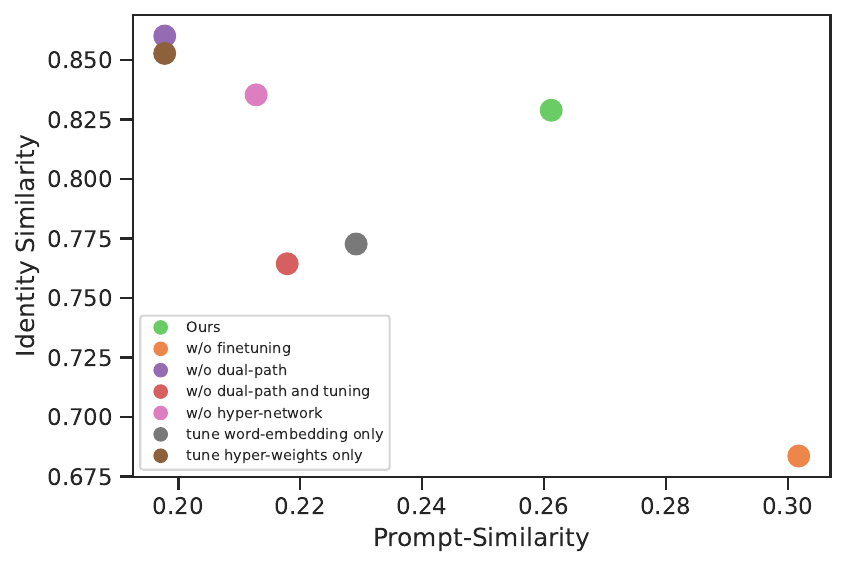}
         \caption{}
         \label{fig:ablation_quant}
     \end{subfigure}

     \caption{Quantitative evaluation results. (a) Comparisons to prior work. Our method presents an appealing point on the identity-prompt similarity trade-off curve, while being orders of magnitude quicker than optimization-based methods. (b) Ablation study results. Removing regularization typically leads to quick overfitting, where editability suffers. Skipping the fine-tuning step harms identity preservation, in line with E4T~\citep{gal2023designing}. }
\end{figure*}

As seen in Fig~\ref{fig:contrastive_qualitative}, incorporating our contrastive-based loss improves results. In particular, omitting any regularization tends to overfit the input image. For example, in the generated image of "A photo of [S*] in the gladiator movie," the word gladiator is overlooked. And the model overfits the predicted token. On the other hand, using our contrastive loss, the generated photo faithfully describes the input prompt while preserving features of the target concept (i.e., the horse). The contrastive loss function also helps to prevent mode collapse by repelling tokens of different images via negative samples. For example, unlike the contrastive-based method, the nearest-neighbor approach does not address mode collapse. It yields less favorable results (See Fig~\ref{fig:contrastive_qualitative}).

\subsection{Comparison with existing methods}
\begin{figure*}[h!]

    \centering
    \includegraphics[width=0.85\linewidth]{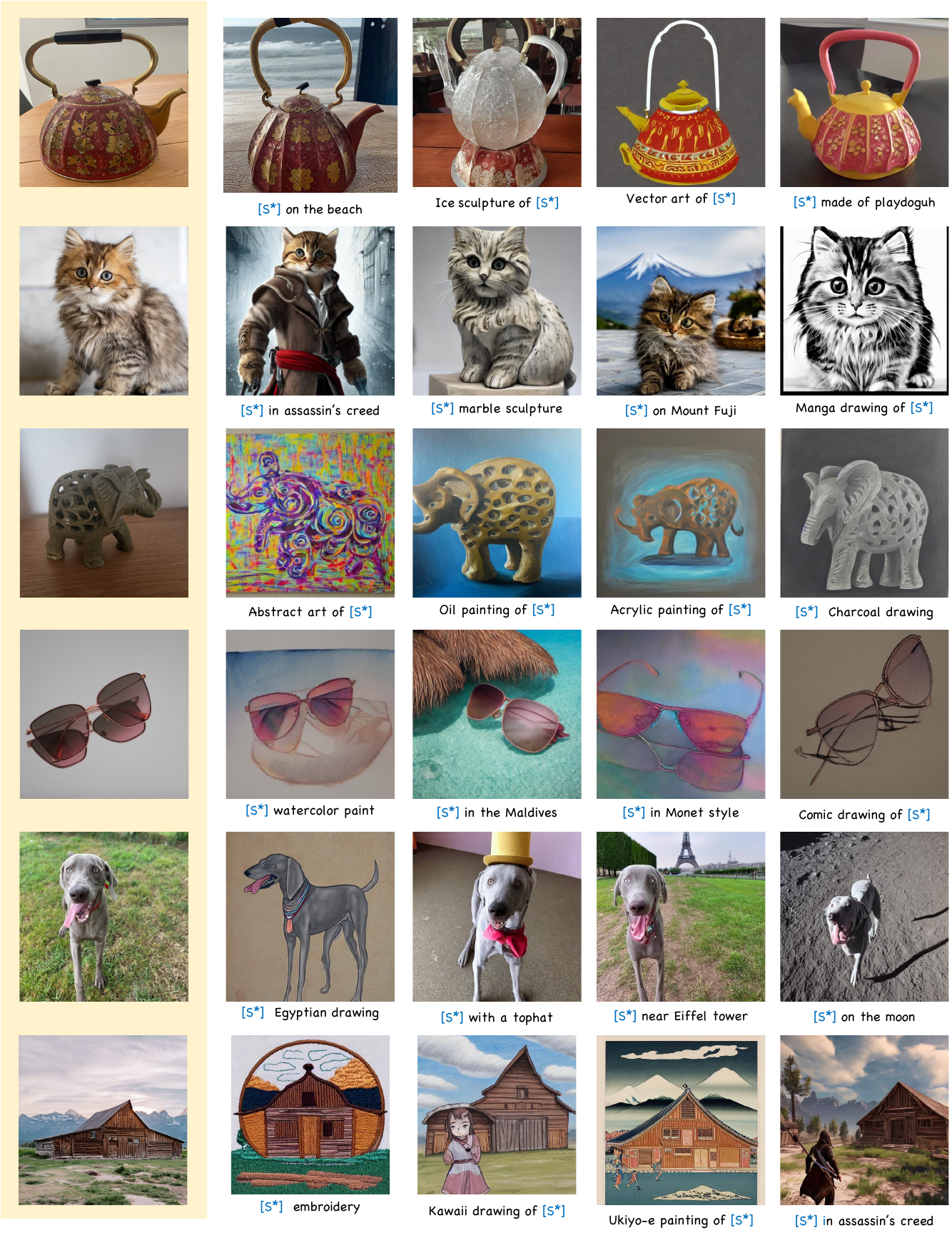}
    \caption{Additional qualitative results generated using our method. The left-most column shows the input image, followed by 4 personalized generations for each subject.}
    \label{fig:additional}
\end{figure*}

We commence our evaluation with a qualitative analysis, demonstrating the ability of our method to capture a remarkable level of detail using a single image and a fraction of the training steps. Figure~\ref{fig:qual_comp} showcases the outcomes of multi-domain personalization by comparing different approaches. Specifically, we compare our method with Textual-Inversion~\cite{gal2022image}, Dream-Booth\cite{ruiz2022dreambooth}, and popular publicly available LoRA library for Stable Diffusion~\cite{simoLoRA2023}. We also compare our method to ELITE~\cite{wei2023elite}, a state-of-the-art multi-domain personalization encoder. For DreamBooth and Textual Inversion, we use the HuggingFace Diffusers implementation~\citep{huggingface2022dreambooth}. Our results are on-par with full tuning-based methods (DreamBooth, LoRA) and significantly outperform the purely-encoder based approach of ELITE, even though the latter has access to additional supervision in the form of segmentation masks. Notably, all tuning-based methods require access to multiple images of the target concept, while our approach utilizes only a single input. Additional results generated using our method can be found in \cref{fig:additional}.

Next, we compare our method to the tuning-based approaches using the CLIP-based metrics. Results are shown in \cref{fig:comparison_quant}. Our method achieves better identity preservation and editability than LoRA, but exhibits a tradeoff when compared to DreamBooth. Notably, it outperforms all baselines when they are trained using only a single image. Overall, our approach is competitive with the state-of-the-art while using only a single image and $12$ tuning iterations.

\subsection{Ablation Analysis}

We conduct an ablation study to better understand the importance of each component in our method. We examine the following setups: removing the dual-path regularization approach, skipping the fine-tuning step, and omitting the hypernetwork branch. We observe that the final tuning step is crucial, inline with the observation from E4T. In particular, when using our baseline without finetuning, we witness a 20\% drop in the object similarity metric. Turning off the dual-path during tuning harms prompt-to-image alignment by nearly 30\%, suggesting heavy overfitting. Hence, we can conclude that the dual-path approach can successfully preserve the prior and diminish overfitting.

Another important component of our method is the hyper-network, which predicts weight modulations to calibrate the generator with our target concept. In our ablation study, we found that omitting the hyper-network at training time negatively impacts the alignment of the generated images with the text prompts. We believe this is because the network must encode more information about the object in the word-embedding, causing attention-overfitting as described in the method sections.

\section{Limitations}
While our approach can extend existing tuning-encoders to multi-class domains, it is still limited by our training data. As such, domains which are poorly represented in the dataset may be hard to encode. As such, a model trained on ImageNet may struggle with cluttered scenes or with human faces.

We believe this limitation can be overcome by training on more general, large-scale datasets such as LAION~\citep{schuhmann2021laion}. However, such an investigation is beyond our resources.

While our method can work across a more general domain, it still requires a tuning-step to increase downstream similarity. However, as the memory requirements and iterations required for such tuning-approaches decreases, they become negligible compared to the time required for synthesis.

\section{Conclusion}

We presented a method for generalizing the tuning-encoder approach beyond a single class domain. Our approach restricts overfitting by ensuring predicted embeddings lie close to the real word domain, and by utilizing a dual-pass approach where the network blends predictions from hard- and soft-prompts. This in turn allows us to quickly personalize a model at inference-time, speeding up personalization by two orders of magnitude compared to optimization-based approaches.

In the future, we hope to further reduce the tuning requirements so that our method can be used on consumer-grade GPUs, allowing end-users to quickly personalize models on their own machine.

\begin{acks}
The first author is supported by the Miriam and Aaron Gutwirth scholarship. This work was partially supported by Len Blavatnik and the Blavatnik family foundation, the Deutsch Foundation, the Yandex Initiative in Machine Learning, BSF (grant 2020280) and ISF (grants 2492/20 and 3441/21). 
\end{acks}

\bibliographystyle{ACM-Reference-Format}
\bibliography{main}


\begin{thebibliography}{74}


\ifx \showCODEN    \undefined \def \showCODEN     #1{\unskip}     \fi
\ifx \showDOI      \undefined \def \showDOI       #1{#1}\fi
\ifx \showISBNx    \undefined \def \showISBNx     #1{\unskip}     \fi
\ifx \showISBNxiii \undefined \def \showISBNxiii  #1{\unskip}     \fi
\ifx \showISSN     \undefined \def \showISSN      #1{\unskip}     \fi
\ifx \showLCCN     \undefined \def \showLCCN      #1{\unskip}     \fi
\ifx \shownote     \undefined \def \shownote      #1{#1}          \fi
\ifx \showarticletitle \undefined \def \showarticletitle #1{#1}   \fi
\ifx \showURL      \undefined \def \showURL       {\relax}        \fi
\providecommand\bibfield[2]{#2}
\providecommand\bibinfo[2]{#2}
\providecommand\natexlab[1]{#1}
\providecommand\showeprint[2][]{arXiv:#2}

\bibitem[sim(2023)]%
        {simoLoRA2023}
 \bibinfo{year}{2023}\natexlab{}.
\newblock \bibinfo{title}{Low-rank Adaptation for Fast Text-to-Image Diffusion
  Fine-tuning}.
\newblock
\newblock


\bibitem[Abdal et~al\mbox{.}(2019)]%
        {abdal2019image2stylegan}
\bibfield{author}{\bibinfo{person}{Rameen Abdal}, \bibinfo{person}{Yipeng Qin},
  {and} \bibinfo{person}{Peter Wonka}.} \bibinfo{year}{2019}\natexlab{}.
\newblock \showarticletitle{Image2stylegan: How to embed images into the
  stylegan latent space?}. In \bibinfo{booktitle}{\emph{Proceedings of the
  IEEE/CVF International Conference on Computer Vision}}.
  \bibinfo{pages}{4432--4441}.
\newblock


\bibitem[Abdal et~al\mbox{.}(2020)]%
        {abdal2020image2stylegan++}
\bibfield{author}{\bibinfo{person}{Rameen Abdal}, \bibinfo{person}{Yipeng Qin},
  {and} \bibinfo{person}{Peter Wonka}.} \bibinfo{year}{2020}\natexlab{}.
\newblock \showarticletitle{Image2stylegan++: How to edit the embedded
  images?}. In \bibinfo{booktitle}{\emph{Proceedings of the IEEE/CVF conference
  on computer vision and pattern recognition}}. \bibinfo{pages}{8296--8305}.
\newblock


\bibitem[Alaluf et~al\mbox{.}(2021)]%
        {alaluf2021hyperstyle}
\bibfield{author}{\bibinfo{person}{Yuval Alaluf}, \bibinfo{person}{Omer Tov},
  \bibinfo{person}{Ron Mokady}, \bibinfo{person}{Rinon Gal}, {and}
  \bibinfo{person}{Amit~H. Bermano}.} \bibinfo{year}{2021}\natexlab{}.
\newblock \bibinfo{title}{HyperStyle: StyleGAN Inversion with HyperNetworks for
  Real Image Editing}.
\newblock
\newblock
\showeprint[arxiv]{2111.15666}~[cs.CV]


\bibitem[Amat et~al\mbox{.}(2018)]%
        {fernando2018artwork}
\bibfield{author}{\bibinfo{person}{Fernando Amat}, \bibinfo{person}{Ashok
  Chandrashekar}, \bibinfo{person}{Tony Jebara}, {and} \bibinfo{person}{Justin
  Basilico}.} \bibinfo{year}{2018}\natexlab{}.
\newblock \showarticletitle{Artwork Personalization at Netflix}. In
  \bibinfo{booktitle}{\emph{Proceedings of the 12th ACM Conference on
  Recommender Systems}} (Vancouver, British Columbia, Canada)
  \emph{(\bibinfo{series}{RecSys '18})}. \bibinfo{publisher}{Association for
  Computing Machinery}, \bibinfo{address}{New York, NY, USA},
  \bibinfo{pages}{487–488}.
\newblock
\showISBNx{9781450359016}
\urldef\tempurl%
\url{https://doi.org/10.1145/3240323.3241729}
\showDOI{\tempurl}


\bibitem[Bai et~al\mbox{.}(2022)]%
        {bai2022high}
\bibfield{author}{\bibinfo{person}{Qingyan Bai}, \bibinfo{person}{Yinghao Xu},
  \bibinfo{person}{Jiapeng Zhu}, \bibinfo{person}{Weihao Xia},
  \bibinfo{person}{Yujiu Yang}, {and} \bibinfo{person}{Yujun Shen}.}
  \bibinfo{year}{2022}\natexlab{}.
\newblock \showarticletitle{High-fidelity GAN inversion with padding space}. In
  \bibinfo{booktitle}{\emph{European Conference on Computer Vision}}. Springer,
  \bibinfo{pages}{36--53}.
\newblock


\bibitem[Balaji et~al\mbox{.}(2022)]%
        {balaji2022ediffi}
\bibfield{author}{\bibinfo{person}{Yogesh Balaji}, \bibinfo{person}{Seungjun
  Nah}, \bibinfo{person}{Xun Huang}, \bibinfo{person}{Arash Vahdat},
  \bibinfo{person}{Jiaming Song}, \bibinfo{person}{Karsten Kreis},
  \bibinfo{person}{Miika Aittala}, \bibinfo{person}{Timo Aila},
  \bibinfo{person}{Samuli Laine}, \bibinfo{person}{Bryan Catanzaro},
  {et~al\mbox{.}}} \bibinfo{year}{2022}\natexlab{}.
\newblock \showarticletitle{ediffi: Text-to-image diffusion models with an
  ensemble of expert denoisers}.
\newblock \bibinfo{journal}{\emph{arXiv preprint arXiv:2211.01324}}
  (\bibinfo{year}{2022}).
\newblock


\bibitem[Bar-Tal et~al\mbox{.}(2022)]%
        {bar2022text2live}
\bibfield{author}{\bibinfo{person}{Omer Bar-Tal}, \bibinfo{person}{Dolev
  Ofri-Amar}, \bibinfo{person}{Rafail Fridman}, \bibinfo{person}{Yoni Kasten},
  {and} \bibinfo{person}{Tali Dekel}.} \bibinfo{year}{2022}\natexlab{}.
\newblock \showarticletitle{Text2LIVE: Text-Driven Layered Image and Video
  Editing}.
\newblock \bibinfo{journal}{\emph{arXiv preprint arXiv:2204.02491}}
  (\bibinfo{year}{2022}).
\newblock


\bibitem[Bau et~al\mbox{.}(2019)]%
        {semantic2019bau}
\bibfield{author}{\bibinfo{person}{David Bau}, \bibinfo{person}{Hendrik
  Strobelt}, \bibinfo{person}{William Peebles}, \bibinfo{person}{Jonas Wulff},
  \bibinfo{person}{Bolei Zhou}, \bibinfo{person}{Jun-Yan Zhu}, {and}
  \bibinfo{person}{Antonio Torralba}.} \bibinfo{year}{2019}\natexlab{}.
\newblock \showarticletitle{Semantic Photo Manipulation with a Generative Image
  Prior}.
\newblock  \bibinfo{volume}{38}, \bibinfo{number}{4} (\bibinfo{year}{2019}).
\newblock
\showISSN{0730-0301}
\urldef\tempurl%
\url{https://doi.org/10.1145/3306346.3323023}
\showDOI{\tempurl}


\bibitem[Benhamdi et~al\mbox{.}(2017)]%
        {benhamdi2017personalized}
\bibfield{author}{\bibinfo{person}{Soulef Benhamdi},
  \bibinfo{person}{Abdesselam Babouri}, {and} \bibinfo{person}{Raja Chiky}.}
  \bibinfo{year}{2017}\natexlab{}.
\newblock \showarticletitle{Personalized recommender system for e-Learning
  environment}.
\newblock \bibinfo{journal}{\emph{Education and Information Technologies}}
  \bibinfo{volume}{22}, \bibinfo{number}{4} (\bibinfo{year}{2017}),
  \bibinfo{pages}{1455--1477}.
\newblock


\bibitem[Brooks et~al\mbox{.}(2023)]%
        {brooks2022instructpix2pix}
\bibfield{author}{\bibinfo{person}{Tim Brooks}, \bibinfo{person}{Aleksander
  Holynski}, {and} \bibinfo{person}{Alexei~A. Efros}.}
  \bibinfo{year}{2023}\natexlab{}.
\newblock \showarticletitle{InstructPix2Pix: Learning to Follow Image Editing
  Instructions}. In \bibinfo{booktitle}{\emph{CVPR}}.
\newblock


\bibitem[Cao et~al\mbox{.}(2022)]%
        {cao2022authentic}
\bibfield{author}{\bibinfo{person}{Chen Cao}, \bibinfo{person}{Tomas Simon},
  \bibinfo{person}{Jin~Kyu Kim}, \bibinfo{person}{Gabe Schwartz},
  \bibinfo{person}{Michael Zollhoefer}, \bibinfo{person}{Shunsuke Saito},
  \bibinfo{person}{Stephen Lombardi}, \bibinfo{person}{Shih-en Wei},
  \bibinfo{person}{Danielle Belko}, \bibinfo{person}{Shoou-i Yu},
  \bibinfo{person}{Yaser Sheikh}, {and} \bibinfo{person}{Jason Saragih}.}
  \bibinfo{year}{2022}\natexlab{}.
\newblock \showarticletitle{Authentic Volumetric Avatars From a Phone Scan}.
\newblock \bibinfo{journal}{\emph{ACM Trans. Graph.}} (\bibinfo{year}{2022}).
\newblock


\bibitem[Chen et~al\mbox{.}(2023)]%
        {Chen2023SubjectdrivenTG}
\bibfield{author}{\bibinfo{person}{Wenhu Chen}, \bibinfo{person}{Hexiang Hu},
  \bibinfo{person}{Yandong Li}, \bibinfo{person}{Nataniel Rui},
  \bibinfo{person}{Xuhui Jia}, \bibinfo{person}{Ming-Wei Chang}, {and}
  \bibinfo{person}{William~W. Cohen}.} \bibinfo{year}{2023}\natexlab{}.
\newblock \showarticletitle{Subject-driven Text-to-Image Generation via
  Apprenticeship Learning}.
\newblock \bibinfo{journal}{\emph{ArXiv}}  \bibinfo{volume}{abs/2304.00186}
  (\bibinfo{year}{2023}).
\newblock


\bibitem[Cho et~al\mbox{.}(2002)]%
        {cho2002personalized}
\bibfield{author}{\bibinfo{person}{Yoon~Ho Cho}, \bibinfo{person}{Jae~Kyeong
  Kim}, {and} \bibinfo{person}{Soung~Hie Kim}.}
  \bibinfo{year}{2002}\natexlab{}.
\newblock \showarticletitle{A personalized recommender system based on web
  usage mining and decision tree induction}.
\newblock \bibinfo{journal}{\emph{Expert systems with Applications}}
  \bibinfo{volume}{23}, \bibinfo{number}{3} (\bibinfo{year}{2002}),
  \bibinfo{pages}{329--342}.
\newblock


\bibitem[Cohen et~al\mbox{.}(2022)]%
        {cohen2022my}
\bibfield{author}{\bibinfo{person}{Niv Cohen}, \bibinfo{person}{Rinon Gal},
  \bibinfo{person}{Eli~A. Meirom}, \bibinfo{person}{Gal Chechik}, {and}
  \bibinfo{person}{Yuval Atzmon}.} \bibinfo{year}{2022}\natexlab{}.
\newblock \showarticletitle{"This is my unicorn, Fluffy": Personalizing frozen
  vision-language representations}. In \bibinfo{booktitle}{\emph{European
  Conference on Computer Vision (ECCV)}}.
\newblock


\bibitem[Dhariwal and Nichol(2021)]%
        {dhariwal2021diffusionBeatsGAN}
\bibfield{author}{\bibinfo{person}{Prafulla Dhariwal} {and}
  \bibinfo{person}{Alexander Nichol}.} \bibinfo{year}{2021}\natexlab{}.
\newblock \showarticletitle{Diffusion models beat gans on image synthesis}.
\newblock \bibinfo{journal}{\emph{Advances in Neural Information Processing
  Systems}}  \bibinfo{volume}{34} (\bibinfo{year}{2021}),
  \bibinfo{pages}{8780--8794}.
\newblock


\bibitem[Dinh et~al\mbox{.}(2022)]%
        {dinh2022hyperinverter}
\bibfield{author}{\bibinfo{person}{Tan~M Dinh}, \bibinfo{person}{Anh~Tuan
  Tran}, \bibinfo{person}{Rang Nguyen}, {and} \bibinfo{person}{Binh-Son Hua}.}
  \bibinfo{year}{2022}\natexlab{}.
\newblock \showarticletitle{Hyperinverter: Improving stylegan inversion via
  hypernetwork}. In \bibinfo{booktitle}{\emph{Proceedings of the IEEE/CVF
  Conference on Computer Vision and Pattern Recognition}}.
  \bibinfo{pages}{11389--11398}.
\newblock


\bibitem[Esser et~al\mbox{.}(2021)]%
        {VQGANEsser}
\bibfield{author}{\bibinfo{person}{Patrick Esser}, \bibinfo{person}{Robin
  Rombach}, {and} \bibinfo{person}{Bj{\"{o}}rn Ommer}.}
  \bibinfo{year}{2021}\natexlab{}.
\newblock \showarticletitle{Taming Transformers for High-Resolution Image
  Synthesis}. In \bibinfo{booktitle}{\emph{{IEEE} Conference on Computer Vision
  and Pattern Recognition, {CVPR} 2021, virtual, June 19-25, 2021}}.
  \bibinfo{publisher}{Computer Vision Foundation / {IEEE}},
  \bibinfo{pages}{12873--12883}.
\newblock
\urldef\tempurl%
\url{https://doi.org/10.1109/CVPR46437.2021.01268}
\showDOI{\tempurl}


\bibitem[Fallah et~al\mbox{.}(2020)]%
        {fallah2020personalized}
\bibfield{author}{\bibinfo{person}{Alireza Fallah}, \bibinfo{person}{Aryan
  Mokhtari}, {and} \bibinfo{person}{Asuman Ozdaglar}.}
  \bibinfo{year}{2020}\natexlab{}.
\newblock \showarticletitle{Personalized federated learning: A meta-learning
  approach}.
\newblock \bibinfo{journal}{\emph{arXiv preprint arXiv:2002.07948}}
  (\bibinfo{year}{2020}).
\newblock


\bibitem[Gal et~al\mbox{.}(2022)]%
        {gal2022image}
\bibfield{author}{\bibinfo{person}{Rinon Gal}, \bibinfo{person}{Yuval Alaluf},
  \bibinfo{person}{Yuval Atzmon}, \bibinfo{person}{Or Patashnik},
  \bibinfo{person}{Amit~H Bermano}, \bibinfo{person}{Gal Chechik}, {and}
  \bibinfo{person}{Daniel Cohen-Or}.} \bibinfo{year}{2022}\natexlab{}.
\newblock \showarticletitle{An image is worth one word: Personalizing
  text-to-image generation using textual inversion}.
\newblock \bibinfo{journal}{\emph{arXiv preprint arXiv:2208.01618}}
  (\bibinfo{year}{2022}).
\newblock


\bibitem[Gal et~al\mbox{.}(2023)]%
        {gal2023designing}
\bibfield{author}{\bibinfo{person}{Rinon Gal}, \bibinfo{person}{Moab Arar},
  \bibinfo{person}{Yuval Atzmon}, \bibinfo{person}{Amit~H Bermano},
  \bibinfo{person}{Gal Chechik}, {and} \bibinfo{person}{Daniel Cohen-Or}.}
  \bibinfo{year}{2023}\natexlab{}.
\newblock \showarticletitle{Designing an encoder for fast personalization of
  text-to-image models}.
\newblock \bibinfo{journal}{\emph{arXiv preprint arXiv:2302.12228}}
  (\bibinfo{year}{2023}).
\newblock


\bibitem[Gal et~al\mbox{.}(2021)]%
        {gal2021stylegan}
\bibfield{author}{\bibinfo{person}{Rinon Gal}, \bibinfo{person}{Or Patashnik},
  \bibinfo{person}{Haggai Maron}, \bibinfo{person}{Gal Chechik}, {and}
  \bibinfo{person}{Daniel Cohen-Or}.} \bibinfo{year}{2021}\natexlab{}.
\newblock \showarticletitle{Stylegan-nada: Clip-guided domain adaptation of
  image generators}.
\newblock \bibinfo{journal}{\emph{arXiv preprint arXiv:2108.00946}}
  (\bibinfo{year}{2021}).
\newblock


\bibitem[Goodfellow et~al\mbox{.}(2014)]%
        {goodfellow2014generative}
\bibfield{author}{\bibinfo{person}{Ian Goodfellow}, \bibinfo{person}{Jean
  Pouget-Abadie}, \bibinfo{person}{Mehdi Mirza}, \bibinfo{person}{Bing Xu},
  \bibinfo{person}{David Warde-Farley}, \bibinfo{person}{Sherjil Ozair},
  \bibinfo{person}{Aaron Courville}, {and} \bibinfo{person}{Yoshua Bengio}.}
  \bibinfo{year}{2014}\natexlab{}.
\newblock \showarticletitle{Generative adversarial nets}.
\newblock \bibinfo{journal}{\emph{Advances in neural information processing
  systems}}  \bibinfo{volume}{27} (\bibinfo{year}{2014}).
\newblock


\bibitem[Gu et~al\mbox{.}(2020)]%
        {gu2020image}
\bibfield{author}{\bibinfo{person}{Jinjin Gu}, \bibinfo{person}{Yujun Shen},
  {and} \bibinfo{person}{Bolei Zhou}.} \bibinfo{year}{2020}\natexlab{}.
\newblock \bibinfo{title}{Image Processing Using Multi-Code GAN Prior}.
\newblock
\newblock
\showeprint[arxiv]{1912.07116}~[cs.CV]


\bibitem[Han et~al\mbox{.}(2023)]%
        {han2023svdiff}
\bibfield{author}{\bibinfo{person}{Ligong Han}, \bibinfo{person}{Yinxiao Li},
  \bibinfo{person}{Han Zhang}, \bibinfo{person}{Peyman Milanfar},
  \bibinfo{person}{Dimitris Metaxas}, {and} \bibinfo{person}{Feng Yang}.}
  \bibinfo{year}{2023}\natexlab{}.
\newblock \bibinfo{title}{SVDiff: Compact Parameter Space for Diffusion
  Fine-Tuning}.
\newblock
\newblock
\showeprint[arxiv]{2303.11305}~[cs.CV]


\bibitem[Hertz et~al\mbox{.}(2022)]%
        {hertz2022prompt}
\bibfield{author}{\bibinfo{person}{Amir Hertz}, \bibinfo{person}{Ron Mokady},
  \bibinfo{person}{Jay Tenenbaum}, \bibinfo{person}{Kfir Aberman},
  \bibinfo{person}{Yael Pritch}, {and} \bibinfo{person}{Daniel Cohen-Or}.}
  \bibinfo{year}{2022}\natexlab{}.
\newblock \showarticletitle{Prompt-to-prompt image editing with cross attention
  control}.
\newblock  (\bibinfo{year}{2022}).
\newblock


\bibitem[Ho et~al\mbox{.}(2020a)]%
        {ho2020denoising}
\bibfield{author}{\bibinfo{person}{Jonathan Ho}, \bibinfo{person}{Ajay Jain},
  {and} \bibinfo{person}{Pieter Abbeel}.} \bibinfo{year}{2020}\natexlab{a}.
\newblock \showarticletitle{Denoising diffusion probabilistic models}.
\newblock \bibinfo{journal}{\emph{Advances in Neural Information Processing
  Systems}}  \bibinfo{volume}{33} (\bibinfo{year}{2020}),
  \bibinfo{pages}{6840--6851}.
\newblock


\bibitem[Ho et~al\mbox{.}(2020b)]%
        {ho2020denoisingDDPM}
\bibfield{author}{\bibinfo{person}{Jonathan Ho}, \bibinfo{person}{Ajay Jain},
  {and} \bibinfo{person}{Pieter Abbeel}.} \bibinfo{year}{2020}\natexlab{b}.
\newblock \showarticletitle{Denoising diffusion probabilistic models}.
\newblock \bibinfo{journal}{\emph{Advances in Neural Information Processing
  Systems}}  \bibinfo{volume}{33} (\bibinfo{year}{2020}),
  \bibinfo{pages}{6840--6851}.
\newblock


\bibitem[Ho and Salimans(2021)]%
        {ho2021classifier}
\bibfield{author}{\bibinfo{person}{Jonathan Ho} {and} \bibinfo{person}{Tim
  Salimans}.} \bibinfo{year}{2021}\natexlab{}.
\newblock \showarticletitle{Classifier-Free Diffusion Guidance}. In
  \bibinfo{booktitle}{\emph{NeurIPS 2021 Workshop on Deep Generative Models and
  Downstream Applications}}.
\newblock


\bibitem[Hu et~al\mbox{.}(2021)]%
        {Hu2021LoRALA}
\bibfield{author}{\bibinfo{person}{Edward~J. Hu}, \bibinfo{person}{Yelong
  Shen}, \bibinfo{person}{Phillip Wallis}, \bibinfo{person}{Zeyuan Allen-Zhu},
  \bibinfo{person}{Yuanzhi Li}, \bibinfo{person}{Shean Wang}, {and}
  \bibinfo{person}{Weizhu Chen}.} \bibinfo{year}{2021}\natexlab{}.
\newblock \showarticletitle{LoRA: Low-Rank Adaptation of Large Language
  Models}.
\newblock \bibinfo{journal}{\emph{ArXiv}}  \bibinfo{volume}{abs/2106.09685}
  (\bibinfo{year}{2021}).
\newblock


\bibitem[Huang et~al\mbox{.}(2023)]%
        {huang2023reversion}
\bibfield{author}{\bibinfo{person}{Ziqi Huang}, \bibinfo{person}{Tianxing Wu},
  \bibinfo{person}{Yuming Jiang}, \bibinfo{person}{Kelvin~CK Chan}, {and}
  \bibinfo{person}{Ziwei Liu}.} \bibinfo{year}{2023}\natexlab{}.
\newblock \showarticletitle{ReVersion: Diffusion-Based Relation Inversion from
  Images}.
\newblock \bibinfo{journal}{\emph{arXiv preprint arXiv:2303.13495}}
  (\bibinfo{year}{2023}).
\newblock


\bibitem[Jiang et~al\mbox{.}(2019)]%
        {jiang2019improving}
\bibfield{author}{\bibinfo{person}{Yihan Jiang}, \bibinfo{person}{Jakub
  Kone{\v{c}}n{\`y}}, \bibinfo{person}{Keith Rush}, {and}
  \bibinfo{person}{Sreeram Kannan}.} \bibinfo{year}{2019}\natexlab{}.
\newblock \showarticletitle{Improving federated learning personalization via
  model agnostic meta learning}.
\newblock \bibinfo{journal}{\emph{arXiv preprint arXiv:1909.12488}}
  (\bibinfo{year}{2019}).
\newblock


\bibitem[Kang et~al\mbox{.}(2023)]%
        {kang2023gigagan}
\bibfield{author}{\bibinfo{person}{Minguk Kang}, \bibinfo{person}{Jun-Yan Zhu},
  \bibinfo{person}{Richard Zhang}, \bibinfo{person}{Jaesik Park},
  \bibinfo{person}{Eli Shechtman}, \bibinfo{person}{Sylvain Paris}, {and}
  \bibinfo{person}{Taesung Park}.} \bibinfo{year}{2023}\natexlab{}.
\newblock \showarticletitle{Scaling up GANs for Text-to-Image Synthesis}. In
  \bibinfo{booktitle}{\emph{Proceedings of the IEEE Conference on Computer
  Vision and Pattern Recognition (CVPR)}}.
\newblock


\bibitem[Kawar et~al\mbox{.}(2022)]%
        {kawar2022imagic}
\bibfield{author}{\bibinfo{person}{Bahjat Kawar}, \bibinfo{person}{Shiran
  Zada}, \bibinfo{person}{Oran Lang}, \bibinfo{person}{Omer Tov},
  \bibinfo{person}{Huiwen Chang}, \bibinfo{person}{Tali Dekel},
  \bibinfo{person}{Inbar Mosseri}, {and} \bibinfo{person}{Michal Irani}.}
  \bibinfo{year}{2022}\natexlab{}.
\newblock \showarticletitle{Imagic: Text-Based Real Image Editing with
  Diffusion Models}.
\newblock \bibinfo{journal}{\emph{arXiv preprint arXiv:2210.09276}}
  (\bibinfo{year}{2022}).
\newblock


\bibitem[Kumari et~al\mbox{.}(2022)]%
        {kumari2022multi}
\bibfield{author}{\bibinfo{person}{Nupur Kumari}, \bibinfo{person}{Bingliang
  Zhang}, \bibinfo{person}{Richard Zhang}, \bibinfo{person}{Eli Shechtman},
  {and} \bibinfo{person}{Jun-Yan Zhu}.} \bibinfo{year}{2022}\natexlab{}.
\newblock \showarticletitle{Multi-Concept Customization of Text-to-Image
  Diffusion}.
\newblock \bibinfo{journal}{\emph{arXiv preprint arXiv:2212.04488}}
  (\bibinfo{year}{2022}).
\newblock


\bibitem[Kuznetsova et~al\mbox{.}(2020)]%
        {OpenImages}
\bibfield{author}{\bibinfo{person}{Alina Kuznetsova}, \bibinfo{person}{Hassan
  Rom}, \bibinfo{person}{Neil Alldrin}, \bibinfo{person}{Jasper Uijlings},
  \bibinfo{person}{Ivan Krasin}, \bibinfo{person}{Jordi Pont-Tuset},
  \bibinfo{person}{Shahab Kamali}, \bibinfo{person}{Stefan Popov},
  \bibinfo{person}{Matteo Malloci}, \bibinfo{person}{Alexander Kolesnikov},
  \bibinfo{person}{Tom Duerig}, {and} \bibinfo{person}{Vittorio Ferrari}.}
  \bibinfo{year}{2020}\natexlab{}.
\newblock \showarticletitle{The Open Images Dataset V4: Unified image
  classification, object detection, and visual relationship detection at
  scale}.
\newblock \bibinfo{journal}{\emph{IJCV}} (\bibinfo{year}{2020}).
\newblock


\bibitem[Li et~al\mbox{.}(2023)]%
        {li2023blipdiffusion}
\bibfield{author}{\bibinfo{person}{Dongxu Li}, \bibinfo{person}{Junnan Li},
  {and} \bibinfo{person}{Steven C.~H. Hoi}.} \bibinfo{year}{2023}\natexlab{}.
\newblock \bibinfo{title}{BLIP-Diffusion: Pre-trained Subject Representation
  for Controllable Text-to-Image Generation and Editing}.
\newblock
\newblock
\showeprint[arxiv]{2305.14720}~[cs.CV]


\bibitem[Mansour et~al\mbox{.}(2020)]%
        {mansour2020three}
\bibfield{author}{\bibinfo{person}{Yishay Mansour}, \bibinfo{person}{Mehryar
  Mohri}, \bibinfo{person}{Jae Ro}, {and} \bibinfo{person}{Ananda~Theertha
  Suresh}.} \bibinfo{year}{2020}\natexlab{}.
\newblock \showarticletitle{Three approaches for personalization with
  applications to federated learning}.
\newblock \bibinfo{journal}{\emph{arXiv preprint arXiv:2002.10619}}
  (\bibinfo{year}{2020}).
\newblock


\bibitem[Martinez et~al\mbox{.}(2009)]%
        {martinez2009s}
\bibfield{author}{\bibinfo{person}{Ana Belen~Barragans Martinez},
  \bibinfo{person}{Jose J~Pazos Arias}, \bibinfo{person}{Ana~Fernandez Vilas},
  \bibinfo{person}{Jorge~Garcia Duque}, {and} \bibinfo{person}{Martin~Lopez
  Nores}.} \bibinfo{year}{2009}\natexlab{}.
\newblock \showarticletitle{What's on TV tonight? An efficient and effective
  personalized recommender system of TV programs}.
\newblock \bibinfo{journal}{\emph{IEEE Transactions on Consumer Electronics}}
  \bibinfo{volume}{55}, \bibinfo{number}{1} (\bibinfo{year}{2009}),
  \bibinfo{pages}{286--294}.
\newblock


\bibitem[Michel et~al\mbox{.}(2021)]%
        {text2mesh}
\bibfield{author}{\bibinfo{person}{Oscar Michel}, \bibinfo{person}{Roi Bar-On},
  \bibinfo{person}{Richard Liu}, \bibinfo{person}{Sagie Benaim}, {and}
  \bibinfo{person}{Rana Hanocka}.} \bibinfo{year}{2021}\natexlab{}.
\newblock \showarticletitle{Text2Mesh: Text-Driven Neural Stylization for
  Meshes}.
\newblock \bibinfo{journal}{\emph{arXiv preprint arXiv:2112.03221}}
  (\bibinfo{year}{2021}).
\newblock


\bibitem[Miech et~al\mbox{.}(2020)]%
        {miech2020endcontrastive}
\bibfield{author}{\bibinfo{person}{Antoine Miech},
  \bibinfo{person}{Jean-Baptiste Alayrac}, \bibinfo{person}{Lucas Smaira},
  \bibinfo{person}{Ivan Laptev}, \bibinfo{person}{Josef Sivic}, {and}
  \bibinfo{person}{Andrew Zisserman}.} \bibinfo{year}{2020}\natexlab{}.
\newblock \showarticletitle{End-to-end learning of visual representations from
  uncurated instructional videos}. In \bibinfo{booktitle}{\emph{Proceedings of
  the IEEE/CVF Conference on Computer Vision and Pattern Recognition}}.
  \bibinfo{pages}{9879--9889}.
\newblock


\bibitem[Mokady et~al\mbox{.}(2022)]%
        {mokady2022null}
\bibfield{author}{\bibinfo{person}{Ron Mokady}, \bibinfo{person}{Amir Hertz},
  \bibinfo{person}{Kfir Aberman}, \bibinfo{person}{Yael Pritch}, {and}
  \bibinfo{person}{Daniel Cohen-Or}.} \bibinfo{year}{2022}\natexlab{}.
\newblock \showarticletitle{Null-text Inversion for Editing Real Images using
  Guided Diffusion Models}.
\newblock \bibinfo{journal}{\emph{arXiv preprint arXiv:2211.09794}}
  (\bibinfo{year}{2022}).
\newblock


\bibitem[Nichol et~al\mbox{.}(2021)]%
        {nichol2021glide}
\bibfield{author}{\bibinfo{person}{Alex Nichol}, \bibinfo{person}{Prafulla
  Dhariwal}, \bibinfo{person}{Aditya Ramesh}, \bibinfo{person}{Pranav Shyam},
  \bibinfo{person}{Pamela Mishkin}, \bibinfo{person}{Bob McGrew},
  \bibinfo{person}{Ilya Sutskever}, {and} \bibinfo{person}{Mark Chen}.}
  \bibinfo{year}{2021}\natexlab{}.
\newblock \showarticletitle{Glide: Towards photorealistic image generation and
  editing with text-guided diffusion models}.
\newblock \bibinfo{journal}{\emph{arXiv preprint arXiv:2112.10741}}
  (\bibinfo{year}{2021}).
\newblock


\bibitem[Nitzan et~al\mbox{.}(2022)]%
        {nitzan2022mystyle}
\bibfield{author}{\bibinfo{person}{Yotam Nitzan}, \bibinfo{person}{Kfir
  Aberman}, \bibinfo{person}{Qiurui He}, \bibinfo{person}{Orly Liba},
  \bibinfo{person}{Michal Yarom}, \bibinfo{person}{Yossi Gandelsman},
  \bibinfo{person}{Inbar Mosseri}, \bibinfo{person}{Yael Pritch}, {and}
  \bibinfo{person}{Daniel Cohen-Or}.} \bibinfo{year}{2022}\natexlab{}.
\newblock \showarticletitle{MyStyle: A Personalized Generative Prior}.
\newblock \bibinfo{journal}{\emph{arXiv preprint arXiv:2203.17272}}
  (\bibinfo{year}{2022}).
\newblock


\bibitem[Parmar et~al\mbox{.}(2022)]%
        {parmar2022spatially}
\bibfield{author}{\bibinfo{person}{Gaurav Parmar}, \bibinfo{person}{Yijun Li},
  \bibinfo{person}{Jingwan Lu}, \bibinfo{person}{Richard Zhang},
  \bibinfo{person}{Jun-Yan Zhu}, {and} \bibinfo{person}{Krishna~Kumar Singh}.}
  \bibinfo{year}{2022}\natexlab{}.
\newblock \showarticletitle{Spatially-Adaptive Multilayer Selection for GAN
  Inversion and Editing}. In \bibinfo{booktitle}{\emph{Proceedings of the
  IEEE/CVF Conference on Computer Vision and Pattern Recognition}}.
  \bibinfo{pages}{11399--11409}.
\newblock


\bibitem[Parmar et~al\mbox{.}(2023)]%
        {parmar2023zeroshot}
\bibfield{author}{\bibinfo{person}{Gaurav Parmar},
  \bibinfo{person}{Krishna~Kumar Singh}, \bibinfo{person}{Richard Zhang},
  \bibinfo{person}{Yijun Li}, \bibinfo{person}{Jingwan Lu}, {and}
  \bibinfo{person}{Jun-Yan Zhu}.} \bibinfo{year}{2023}\natexlab{}.
\newblock \bibinfo{title}{Zero-shot Image-to-Image Translation}.
\newblock
\newblock
\showeprint[arxiv]{2302.03027}~[cs.CV]


\bibitem[Patashnik et~al\mbox{.}(2021)]%
        {patashnik2021styleclip}
\bibfield{author}{\bibinfo{person}{Or Patashnik}, \bibinfo{person}{Zongze Wu},
  \bibinfo{person}{Eli Shechtman}, \bibinfo{person}{Daniel Cohen-Or}, {and}
  \bibinfo{person}{Dani Lischinski}.} \bibinfo{year}{2021}\natexlab{}.
\newblock \showarticletitle{StyleCLIP: Text-Driven Manipulation of StyleGAN
  Imagery}.
\newblock \bibinfo{journal}{\emph{arXiv preprint arXiv:2103.17249}}
  (\bibinfo{year}{2021}).
\newblock


\bibitem[Patil and Cuenca(2022)]%
        {huggingface2022dreambooth}
\bibfield{author}{\bibinfo{person}{Suraj Patil} {and} \bibinfo{person}{Pedro
  Cuenca}.} \bibinfo{year}{2022}\natexlab{}.
\newblock \bibinfo{title}{HuggingFace DreamBooth Implementation}.
\newblock
  \bibinfo{howpublished}{\url{https://huggingface.co/docs/diffusers/training/dreambooth}}.
\newblock


\bibitem[Pidhorskyi et~al\mbox{.}(2020)]%
        {pidhorskyi2020adversarial}
\bibfield{author}{\bibinfo{person}{Stanislav Pidhorskyi},
  \bibinfo{person}{Donald~A Adjeroh}, {and} \bibinfo{person}{Gianfranco
  Doretto}.} \bibinfo{year}{2020}\natexlab{}.
\newblock \showarticletitle{Adversarial Latent Autoencoders}. In
  \bibinfo{booktitle}{\emph{Proceedings of the IEEE/CVF Conference on Computer
  Vision and Pattern Recognition}}. \bibinfo{pages}{14104--14113}.
\newblock


\bibitem[Radford et~al\mbox{.}(2021)]%
        {radford2021learning}
\bibfield{author}{\bibinfo{person}{Alec Radford}, \bibinfo{person}{Jong~Wook
  Kim}, \bibinfo{person}{Chris Hallacy}, \bibinfo{person}{Aditya Ramesh},
  \bibinfo{person}{Gabriel Goh}, \bibinfo{person}{Sandhini Agarwal},
  \bibinfo{person}{Girish Sastry}, \bibinfo{person}{Amanda Askell},
  \bibinfo{person}{Pamela Mishkin}, \bibinfo{person}{Jack Clark},
  {et~al\mbox{.}}} \bibinfo{year}{2021}\natexlab{}.
\newblock \showarticletitle{Learning transferable visual models from natural
  language supervision}.
\newblock \bibinfo{journal}{\emph{arXiv preprint arXiv:2103.00020}}
  (\bibinfo{year}{2021}).
\newblock


\bibitem[Ramesh et~al\mbox{.}(2022)]%
        {ramesh2022hierarchical}
\bibfield{author}{\bibinfo{person}{Aditya Ramesh}, \bibinfo{person}{Prafulla
  Dhariwal}, \bibinfo{person}{Alex Nichol}, \bibinfo{person}{Casey Chu}, {and}
  \bibinfo{person}{Mark Chen}.} \bibinfo{year}{2022}\natexlab{}.
\newblock \showarticletitle{Hierarchical text-conditional image generation with
  clip latents}.
\newblock \bibinfo{journal}{\emph{arXiv preprint arXiv:2204.06125}}
  (\bibinfo{year}{2022}).
\newblock


\bibitem[Richardson et~al\mbox{.}(2020)]%
        {richardson2020encoding}
\bibfield{author}{\bibinfo{person}{Elad Richardson}, \bibinfo{person}{Yuval
  Alaluf}, \bibinfo{person}{Or Patashnik}, \bibinfo{person}{Yotam Nitzan},
  \bibinfo{person}{Yaniv Azar}, \bibinfo{person}{Stav Shapiro}, {and}
  \bibinfo{person}{Daniel Cohen-Or}.} \bibinfo{year}{2020}\natexlab{}.
\newblock \showarticletitle{Encoding in Style: a StyleGAN Encoder for
  Image-to-Image Translation}.
\newblock \bibinfo{journal}{\emph{arXiv preprint arXiv:2008.00951}}
  (\bibinfo{year}{2020}).
\newblock


\bibitem[Richardson et~al\mbox{.}(2023)]%
        {richardson2023texture}
\bibfield{author}{\bibinfo{person}{Elad Richardson}, \bibinfo{person}{Gal
  Metzer}, \bibinfo{person}{Yuval Alaluf}, \bibinfo{person}{Raja Giryes}, {and}
  \bibinfo{person}{Daniel Cohen-Or}.} \bibinfo{year}{2023}\natexlab{}.
\newblock \showarticletitle{Texture: Text-guided texturing of 3d shapes}.
\newblock \bibinfo{journal}{\emph{arXiv preprint arXiv:2302.01721}}
  (\bibinfo{year}{2023}).
\newblock


\bibitem[Roich et~al\mbox{.}(2021)]%
        {roich2021pivotal}
\bibfield{author}{\bibinfo{person}{Daniel Roich}, \bibinfo{person}{Ron Mokady},
  \bibinfo{person}{Amit~H Bermano}, {and} \bibinfo{person}{Daniel Cohen-Or}.}
  \bibinfo{year}{2021}\natexlab{}.
\newblock \showarticletitle{Pivotal tuning for latent-based editing of real
  images}.
\newblock \bibinfo{journal}{\emph{arXiv preprint arXiv:2106.05744}}
  (\bibinfo{year}{2021}).
\newblock


\bibitem[Rombach et~al\mbox{.}(2022)]%
        {rombach2021highresolutionLDM}
\bibfield{author}{\bibinfo{person}{Robin Rombach}, \bibinfo{person}{Andreas
  Blattmann}, \bibinfo{person}{Dominik Lorenz}, \bibinfo{person}{Patrick
  Esser}, {and} \bibinfo{person}{Bj{\"{o}}rn Ommer}.}
  \bibinfo{year}{2022}\natexlab{}.
\newblock \showarticletitle{High-Resolution Image Synthesis with Latent
  Diffusion Models}. In \bibinfo{booktitle}{\emph{{IEEE/CVF} Conference on
  Computer Vision and Pattern Recognition, {CVPR} 2022, New Orleans, LA, USA,
  June 18-24, 2022}}. \bibinfo{publisher}{{IEEE}},
  \bibinfo{pages}{10674--10685}.
\newblock
\urldef\tempurl%
\url{https://doi.org/10.1109/CVPR52688.2022.01042}
\showDOI{\tempurl}


\bibitem[Ruiz et~al\mbox{.}(2022)]%
        {ruiz2022dreambooth}
\bibfield{author}{\bibinfo{person}{Nataniel Ruiz}, \bibinfo{person}{Yuanzhen
  Li}, \bibinfo{person}{Varun Jampani}, \bibinfo{person}{Yael Pritch},
  \bibinfo{person}{Michael Rubinstein}, {and} \bibinfo{person}{Kfir Aberman}.}
  \bibinfo{year}{2022}\natexlab{}.
\newblock \showarticletitle{DreamBooth: Fine Tuning Text-to-image Diffusion
  Models for Subject-Driven Generation}.
\newblock  (\bibinfo{year}{2022}).
\newblock


\bibitem[Russakovsky et~al\mbox{.}(2015)]%
        {ILSVRC15}
\bibfield{author}{\bibinfo{person}{Olga Russakovsky}, \bibinfo{person}{Jia
  Deng}, \bibinfo{person}{Hao Su}, \bibinfo{person}{Jonathan Krause},
  \bibinfo{person}{Sanjeev Satheesh}, \bibinfo{person}{Sean Ma},
  \bibinfo{person}{Zhiheng Huang}, \bibinfo{person}{Andrej Karpathy},
  \bibinfo{person}{Aditya Khosla}, \bibinfo{person}{Michael Bernstein},
  \bibinfo{person}{Alexander~C. Berg}, {and} \bibinfo{person}{Li Fei-Fei}.}
  \bibinfo{year}{2015}\natexlab{}.
\newblock \showarticletitle{{ImageNet Large Scale Visual Recognition
  Challenge}}.
\newblock \bibinfo{journal}{\emph{International Journal of Computer Vision
  (IJCV)}} \bibinfo{volume}{115}, \bibinfo{number}{3} (\bibinfo{year}{2015}),
  \bibinfo{pages}{211--252}.
\newblock
\urldef\tempurl%
\url{https://doi.org/10.1007/s11263-015-0816-y}
\showDOI{\tempurl}


\bibitem[Saharia et~al\mbox{.}(2022)]%
        {saharia2022photorealistic}
\bibfield{author}{\bibinfo{person}{Chitwan Saharia}, \bibinfo{person}{William
  Chan}, \bibinfo{person}{Saurabh Saxena}, \bibinfo{person}{Lala Li},
  \bibinfo{person}{Jay Whang}, \bibinfo{person}{Emily Denton},
  \bibinfo{person}{Seyed Kamyar~Seyed Ghasemipour},
  \bibinfo{person}{Burcu~Karagol Ayan}, \bibinfo{person}{S~Sara Mahdavi},
  \bibinfo{person}{Rapha~Gontijo Lopes}, {et~al\mbox{.}}}
  \bibinfo{year}{2022}\natexlab{}.
\newblock \showarticletitle{Photorealistic Text-to-Image Diffusion Models with
  Deep Language Understanding}.
\newblock \bibinfo{journal}{\emph{arXiv preprint arXiv:2205.11487}}
  (\bibinfo{year}{2022}).
\newblock


\bibitem[Sauer et~al\mbox{.}(2023)]%
        {Sauer2023ICML}
\bibfield{author}{\bibinfo{person}{Axel Sauer}, \bibinfo{person}{Tero Karras},
  \bibinfo{person}{Samuli Laine}, \bibinfo{person}{Andreas Geiger}, {and}
  \bibinfo{person}{Timo Aila}.} \bibinfo{year}{2023}\natexlab{}.
\newblock \showarticletitle{{StyleGAN-T}: Unlocking the Power of {GANs} for
  Fast Large-Scale Text-to-Image Synthesis}.
\newblock \bibinfo{journal}{\emph{International Conference on Machine
  Learning}}  \bibinfo{volume}{abs/2301.09515}.
\newblock
\urldef\tempurl%
\url{https://arxiv.org/abs/2301.09515}
\showURL{%
\tempurl}


\bibitem[Schuhmann et~al\mbox{.}(2021)]%
        {schuhmann2021laion}
\bibfield{author}{\bibinfo{person}{Christoph Schuhmann},
  \bibinfo{person}{Richard Vencu}, \bibinfo{person}{Romain Beaumont},
  \bibinfo{person}{Robert Kaczmarczyk}, \bibinfo{person}{Clayton Mullis},
  \bibinfo{person}{Aarush Katta}, \bibinfo{person}{Theo Coombes},
  \bibinfo{person}{Jenia Jitsev}, {and} \bibinfo{person}{Aran Komatsuzaki}.}
  \bibinfo{year}{2021}\natexlab{}.
\newblock \showarticletitle{Laion-400m: Open dataset of clip-filtered 400
  million image-text pairs}.
\newblock \bibinfo{journal}{\emph{arXiv preprint arXiv:2111.02114}}
  (\bibinfo{year}{2021}).
\newblock


\bibitem[Shamsian et~al\mbox{.}(2021)]%
        {shamsian2021personalized}
\bibfield{author}{\bibinfo{person}{Aviv Shamsian}, \bibinfo{person}{Aviv
  Navon}, \bibinfo{person}{Ethan Fetaya}, {and} \bibinfo{person}{Gal Chechik}.}
  \bibinfo{year}{2021}\natexlab{}.
\newblock \showarticletitle{Personalized federated learning using
  hypernetworks}. In \bibinfo{booktitle}{\emph{International Conference on
  Machine Learning}}. PMLR, \bibinfo{pages}{9489--9502}.
\newblock


\bibitem[Shi et~al\mbox{.}(2023)]%
        {shi2023instantbooth}
\bibfield{author}{\bibinfo{person}{Jing Shi}, \bibinfo{person}{Wei Xiong},
  \bibinfo{person}{Zhe Lin}, {and} \bibinfo{person}{Hyun~Joon Jung}.}
  \bibinfo{year}{2023}\natexlab{}.
\newblock \bibinfo{title}{InstantBooth: Personalized Text-to-Image Generation
  without Test-Time Finetuning}.
\newblock
\newblock
\showeprint[arxiv]{2304.03411}~[cs.CV]


\bibitem[Song et~al\mbox{.}(2020)]%
        {song2020denoising}
\bibfield{author}{\bibinfo{person}{Jiaming Song}, \bibinfo{person}{Chenlin
  Meng}, {and} \bibinfo{person}{Stefano Ermon}.}
  \bibinfo{year}{2020}\natexlab{}.
\newblock \showarticletitle{Denoising Diffusion Implicit Models}. In
  \bibinfo{booktitle}{\emph{International Conference on Learning
  Representations}}.
\newblock


\bibitem[Tewel et~al\mbox{.}(2023)]%
        {tewel2023key}
\bibfield{author}{\bibinfo{person}{Yoad Tewel}, \bibinfo{person}{Rinon Gal},
  \bibinfo{person}{Gal Chechik}, {and} \bibinfo{person}{Yuval Atzmon}.}
  \bibinfo{year}{2023}\natexlab{}.
\newblock \showarticletitle{Key-Locked Rank One Editing for Text-to-Image
  Personalization}.
\newblock \bibinfo{journal}{\emph{arXiv preprint arXiv:2305.01644}}
  (\bibinfo{year}{2023}).
\newblock


\bibitem[Tov et~al\mbox{.}(2021)]%
        {tov2021designing}
\bibfield{author}{\bibinfo{person}{Omer Tov}, \bibinfo{person}{Yuval Alaluf},
  \bibinfo{person}{Yotam Nitzan}, \bibinfo{person}{Or Patashnik}, {and}
  \bibinfo{person}{Daniel Cohen-Or}.} \bibinfo{year}{2021}\natexlab{}.
\newblock \showarticletitle{Designing an Encoder for StyleGAN Image
  Manipulation}.
\newblock \bibinfo{journal}{\emph{arXiv preprint arXiv:2102.02766}}
  (\bibinfo{year}{2021}).
\newblock


\bibitem[Tumanyan et~al\mbox{.}(2022)]%
        {tumanyan2022plug}
\bibfield{author}{\bibinfo{person}{Narek Tumanyan}, \bibinfo{person}{Michal
  Geyer}, \bibinfo{person}{Shai Bagon}, {and} \bibinfo{person}{Tali Dekel}.}
  \bibinfo{year}{2022}\natexlab{}.
\newblock \showarticletitle{Plug-and-Play Diffusion Features for Text-Driven
  Image-to-Image Translation}.
\newblock \bibinfo{journal}{\emph{arXiv preprint arXiv:2211.12572}}
  (\bibinfo{year}{2022}).
\newblock


\bibitem[Voynov et~al\mbox{.}(2023)]%
        {voynov2023p+}
\bibfield{author}{\bibinfo{person}{Andrey Voynov}, \bibinfo{person}{Qinghao
  Chu}, \bibinfo{person}{Daniel Cohen-Or}, {and} \bibinfo{person}{Kfir
  Aberman}.} \bibinfo{year}{2023}\natexlab{}.
\newblock \showarticletitle{$ P+ $: Extended Textual Conditioning in
  Text-to-Image Generation}.
\newblock \bibinfo{journal}{\emph{arXiv preprint arXiv:2303.09522}}
  (\bibinfo{year}{2023}).
\newblock


\bibitem[Wang et~al\mbox{.}(2022)]%
        {wang2021HFGI}
\bibfield{author}{\bibinfo{person}{Tengfei Wang}, \bibinfo{person}{Yong Zhang},
  \bibinfo{person}{Yanbo Fan}, \bibinfo{person}{Jue Wang}, {and}
  \bibinfo{person}{Qifeng Chen}.} \bibinfo{year}{2022}\natexlab{}.
\newblock \showarticletitle{High-Fidelity GAN Inversion for Image Attribute
  Editing}. In \bibinfo{booktitle}{\emph{Proceedings of the IEEE/CVF Conference
  on Computer Vision and Pattern Recognition (CVPR)}}.
\newblock


\bibitem[Wei et~al\mbox{.}(2023)]%
        {wei2023elite}
\bibfield{author}{\bibinfo{person}{Yuxiang Wei}, \bibinfo{person}{Yabo Zhang},
  \bibinfo{person}{Zhilong Ji}, \bibinfo{person}{Jinfeng Bai},
  \bibinfo{person}{Lei Zhang}, {and} \bibinfo{person}{Wangmeng Zuo}.}
  \bibinfo{year}{2023}\natexlab{}.
\newblock \showarticletitle{Elite: Encoding visual concepts into textual
  embeddings for customized text-to-image generation}.
\newblock \bibinfo{journal}{\emph{arXiv preprint arXiv:2302.13848}}
  (\bibinfo{year}{2023}).
\newblock


\bibitem[Xia et~al\mbox{.}(2021)]%
        {xia2021gan}
\bibfield{author}{\bibinfo{person}{Weihao Xia}, \bibinfo{person}{Yulun Zhang},
  \bibinfo{person}{Yujiu Yang}, \bibinfo{person}{Jing-Hao Xue},
  \bibinfo{person}{Bolei Zhou}, {and} \bibinfo{person}{Ming-Hsuan Yang}.}
  \bibinfo{year}{2021}\natexlab{}.
\newblock \bibinfo{title}{GAN Inversion: A Survey}.
\newblock
\newblock
\showeprint[arxiv]{2101.05278}~[cs.CV]


\bibitem[Zhou et~al\mbox{.}(2023)]%
        {zhou2023enhancing}
\bibfield{author}{\bibinfo{person}{Yufan Zhou}, \bibinfo{person}{Ruiyi Zhang},
  \bibinfo{person}{Tong Sun}, {and} \bibinfo{person}{Jinhui Xu}.}
  \bibinfo{year}{2023}\natexlab{}.
\newblock \bibinfo{title}{Enhancing Detail Preservation for Customized
  Text-to-Image Generation: A Regularization-Free Approach}.
\newblock
\newblock
\showeprint[arxiv]{2305.13579}~[cs.CV]


\bibitem[Zhu et~al\mbox{.}(2020b)]%
        {zhu2020domain}
\bibfield{author}{\bibinfo{person}{Jiapeng Zhu}, \bibinfo{person}{Yujun Shen},
  \bibinfo{person}{Deli Zhao}, {and} \bibinfo{person}{Bolei Zhou}.}
  \bibinfo{year}{2020}\natexlab{b}.
\newblock \showarticletitle{In-domain gan inversion for real image editing}.
\newblock \bibinfo{journal}{\emph{arXiv preprint arXiv:2004.00049}}
  (\bibinfo{year}{2020}).
\newblock


\bibitem[Zhu et~al\mbox{.}(2016)]%
        {zhu2016generative}
\bibfield{author}{\bibinfo{person}{Jun-Yan Zhu}, \bibinfo{person}{Philipp
  Kr{\"a}henb{\"u}hl}, \bibinfo{person}{Eli Shechtman}, {and}
  \bibinfo{person}{Alexei~A Efros}.} \bibinfo{year}{2016}\natexlab{}.
\newblock \showarticletitle{Generative visual manipulation on the natural image
  manifold}. In \bibinfo{booktitle}{\emph{European conference on computer
  vision}}. Springer, \bibinfo{pages}{597--613}.
\newblock


\bibitem[Zhu et~al\mbox{.}(2020a)]%
        {zhu2020improved}
\bibfield{author}{\bibinfo{person}{Peihao Zhu}, \bibinfo{person}{Rameen Abdal},
  \bibinfo{person}{Yipeng Qin}, {and} \bibinfo{person}{Peter Wonka}.}
  \bibinfo{year}{2020}\natexlab{a}.
\newblock \bibinfo{title}{Improved StyleGAN Embedding: Where are the Good
  Latents?}
\newblock
\newblock
\showeprint[arxiv]{2012.09036}~[cs.CV]


\end{thebibliography}

\clearpage

\end{document}